\documentclass[11pt]{article}

\usepackage[final]{acl}

\usepackage{times}
\usepackage{latexsym}

\usepackage[T1]{fontenc}

\usepackage[utf8]{inputenc}

\usepackage{microtype}

\usepackage{inconsolata}

\usepackage{graphicx}

\usepackage{times,latexsym}
\usepackage{url}
\usepackage[T1]{fontenc}
\usepackage{amsmath,amssymb,amsfonts}
\usepackage{graphicx}
\usepackage{xcolor}
\usepackage{tikz}
\usepackage{pgfplots}
\usepackage{booktabs}
\usepackage{multirow}
\usepackage{subcaption}
\usepackage{algorithm}
\usepackage{algorithmic}
\usepackage{pifont}
\newcommand{\cmark}{\ding{51}}
\newcommand{\xmark}{\ding{55}}

\usetikzlibrary{shapes,arrows,positioning,fit,backgrounds}

\title{A Layer-wise Analysis of Supervised Fine-Tuning}

\author{
  Qinghua Zhao\textsuperscript{1}, 
  Xueling Gong\textsuperscript{1}, 
  Xinyu Chen\textsuperscript{1}, 
  Zhongfeng Kang\textsuperscript{2}, 
  Xinlu Li\textsuperscript{1}\thanks{~~Corresponding author.} \\
  \textsuperscript{1}Hefei University \quad
  \textsuperscript{2}Lanzhou University \\
  \texttt{\{zhaoqh, xinlu.li\}@hfuu.edu.cn} \quad
  \texttt{kangzf@lzu.edu.cn} \\
  \texttt{\{gongxueling, chenxinyu\}@stu.hfuu.edu.cn}
}

\begin{document}
\maketitle
\begin{abstract}
While critical for alignment, Supervised Fine-Tuning (SFT) incurs the risk of catastrophic forgetting, yet the layer-wise emergence of instruction-following capabilities remains elusive. We investigate this mechanism via a comprehensive analysis utilizing information-theoretic, geometric, and optimization metrics across model scales (1B-32B). Our experiments reveal a distinct depth-dependent pattern: middle layers (20\%-80\%) are stable, whereas final layers exhibit high sensitivity.  Leveraging this insight, we propose Mid-Block Efficient Tuning, which selectively updates these critical intermediate layers. Empirically, our method outperforms standard LoRA up to 10.2\% on GSM8K (OLMo2-7B) with reduced parameter overhead, demonstrating that effective alignment is architecturally localized rather than distributed. The code is publicly available at \url{https://github.com/lshowway/base}.
\end{abstract}

\section{Introduction}
Supervised Fine-Tuning has established itself as the cornerstone for aligning Large Language Models (LLMs) with human intent. Its efficacy is underscored by the observation that minimal supervision, as few as 1,000 curated examples, suffices to drastically transform base models into capable instruction-following agents~\cite{lima2023, ouyang2022training,agenteaicde25,hydra26}.

Despite this empirical success, the mechanisms driving SFT remain nuanced. Research indicates that SFT primarily recalibrates attention patterns and shifts stylistic token distributions rather than altering underlying knowledge, effectively functioning as a ``surface-level'' adaptation~\cite{wu-etal-2024-language, lin2024the, wei2022finetuned}. However, current parameter-efficient fine-tuning  methods like LoRA ignore this depth-dependent heterogeneity. By applying updates {uniformly across all layers}~\cite{pmlr-v235-ghosh24a}, these methods operate under the suboptimal assumption that all layers contribute equally to alignment, potentially wasting parameter budget on insensitive layers.

A critical gap remains: while we know {what} changes during supervised fine-tuning, we have limited insight into {where these changes occur across the model's depth} and {which layers are essential} for instruction-following capabilities. Prior work has explored layer-wise knowledge localization \cite{meng2022locating} and layer-wise probing \cite{tenney2019bert}, but these studies focused on {what knowledge is stored where} rather than {where task adaptation occurs during fine-tuning}. Understanding these layer-specific dynamics is crucial not only for theoretical insights but also for developing more efficient alignment procedures that concentrate computational resources where they matter most.

We  conduct layer-wise analysis across models from 1B to 32B parameters, employing information-theoretic (entropy, effective rank), geometric (CKA, cosine similarity), and optimization (weight change) metrics. Through  layer-wise probing, weight change tracking, and layer swapping and selective LoRA fine-tuning, we uncover a consistent {depth-dependent adaptation pattern}: 1) Representation similarity between Base and SFT models exhibits a progressive decline, culminating in a {precipitous drop} in the final layers; 2) Internal representations within both Base and SFT models undergo {drastic divergence} specifically in the upper layers; 3) Parameter update magnitudes mirror this trajectory, exhibiting significantly {higher intensity} in the top layers.

Based on this finding, we propose {Mid-Block Efficient Tuning}, which selectively updates only these critical intermediate layers. Experiments on mathematical reasoning (GSM8K) demonstrate that mid-block tuning achieves 37.5\% accuracy, a 10-percentage-point improvement over standard LoRA (28\%). The tendency is consistent across OLMo2-7B, 13B, 32B, and Mistral-7B, suggesting the pattern generalizes across model architectures and scales. Comparative studies confirm that targeting edge layers (bottom 20\% or top 20\%) results in  performance degradation, validating the architectural locality of effective alignment.
It is important to note that Mid-Block Efficient Tuning is not intended as a competing alternative to existing PEFT methods such as QLoRA or AdaLoRA. Rather, it serves as an analysis-driven proof-of-concept that validates our mechanistic findings regarding depth-dependent adaptation. We deliberately adopt standard LoRA as our primary baseline to isolate the effect of layer depth selection.

Drawing on these findings, we posit that supervised fine-tuning shares the same fundamental optimization dynamics as pre-training, i.e., updating parameters via backpropagation to encode new information, yet differs critically in data scale. We identify a functional divergence driven by these dynamics: the aggressive plasticity in the top layers causes incoming information to overwrite pre-existing features, marking these layers as the primary locus of catastrophic forgetting. Conversely, new information integrates with prior knowledge within the intermediate layers, which serve as the stable substrate for memory consolidation.

\section{{Related Work}}
We categorize related work into applications demonstrating the efficacy of SFT, and analytical studies probing the internal mechanisms behind these improvements.

\subsection{Understanding SFT Effects}
SFT functions as a critical mechanism for instruction alignment and generalization. Its efficacy is demonstrated across disparate tasks, including mathematical problem-solving~\cite{wei2022finetuned, 10.5555/3722577.3722647, 10.5555/3692070.3694024}, visual-language understanding~\cite{NEURIPS2023_6dcf277e}, question generation~\cite{agentqgcikm24,kgqgipm24}, and  suppressing toxic outputs and mitigating social biases~\cite{10.1145/3630106.3658979}. 

However, SFT is constrained by data scarcity and knowledge boundaries, often incurring an ``alignment tax'' ~\citep{ouyang2022training} where general capability gains are offset by catastrophic forgetting in specialized reasoning and increased hallucinations~\cite{jiang2025unlocking, pmlr-v235-ghosh24a}. Beyond these performance trade-offs, the safety impact of SFT remains nuanced: rather than monotonically improving alignment, the process can paradoxically compromise pre-training guardrails or even exacerbate social biases relative to base models~\cite{lyu2024keeping, 10.1162/tacl_a_00673}.

\subsection{Interpretability of Fine-tuning}
A dominant perspective, the \textit{Surface Alignment Hypothesis}, posits that SFT primarily elicits pre-trained capabilities rather than injecting new knowledge. This view is supported by observations that minimal data (e.g., 1K samples) suffices for robust alignment, suggesting SFT acts merely as a stylistic steer towards user-preferred formats~\cite{lima2023, jha2023limitinstructiontuningevaluation, kung-peng-2023-models}. Such findings are corroborated by distributional analyses, which reveal that SFT induces shifts predominantly in stylistic tokens while preserving semantic representations~\cite{lin2024the}.

From a mechanistic perspective, SFT recalibrates model behavior by sharpening attention on instruction tokens~\cite{wu-etal-2024-language} and inducing representations in which closer layers between the base and SFT models share higher similarity~\cite{rimsky-etal-2024-steering}. While this process enhances alignment with human cognition~\cite{aw2024instructiontuning}, it respects inherent knowledge boundaries: SFT optimizes response confidence rather than altering factual storage locations~\cite{ren2025learning, du2025posttrainingreshapesllmsmechanistic}. Consequently, forcing knowledge injection during this stage disrupts internal consistency and exacerbates hallucinations~\cite{ren-etal-2024-learning, pmlr-v235-ghosh24a}. Furthermore, feature localization via sparse autoencoders identifies the final layers as the critical locus for driving instruction adherence~\cite{he2025saifsparseautoencoderframework}.

\section{{Methodology}}
\subsection{Preliminaries}
To deconstruct the layer-wise evolutionary dynamics induced by supervised fine-tuning, we first establish the notation for model representations and the spectral analysis framework. We analyze the internal representation space of a Base model $\mathcal{M}_{b}$ and its SFT counterpart $\mathcal{M}_{s}$. Both models share an identical architecture with $L$ layers. {Let $h^{(l)} \in \mathbb{R}^{T \times D}$ }denote the hidden states at layer $l$ for an input sequence of length $T$, where $D$ is the hidden dimension. Given a dataset $\mathcal{D}=\{x_{i}\}_{i=1}^{N}$, we construct two distinct types of representation matrices to capture both local and global dynamics. First, the \textit{{token-level}} $H_{i}^{(l)} \in \mathbb{R}^{T \times D}$ is composed of the representations of all tokens within the $i$-th sample, which preserves fine-grained sequential details. Second, the \textit{{dataset-level}} $\overline{H}^{(l)} \in \mathbb{R}^{N \times D}$ is constructed by stacking the mean-pooled vectors $\tilde{h}_{i}^{(l)} = \frac{1}{T}\sum_{t=1}^{T} h_{i}^{(l)}[t]$ from all $N$ samples, representing the global data manifold.

Our analytical framework is grounded in the spectral properties of the Gram Matrix. For a generic representation matrix $Z \in \mathbb{R}^{T \times D}$ for a token-level $H_{i}^{(l)}$ or  $Z \in \mathbb{R}^{N \times D}$ a dataset-level $\overline{H}^{(l)}$), the Gram matrix is defined as {$K = Z Z^{\top}$, where the entry $K_{jk}$ captures the pairwise similarity between elements.} To quantify the intrinsic information capacity and dimensionality of the representation space, we employ the $\alpha$-order matrix-based entropy $S_\alpha(K) = \frac{1}{1-\alpha} \log_2 \text{tr} \left( \left( \frac{K}{\text{tr}(K)} \right)^\alpha \right) = \frac{1}{1-\alpha} \log_2 \left( \sum_{j} \tilde{\lambda}_{j}^\alpha \right)$,
where $\tilde{\lambda}_{j}$ are the eigenvalues of the normalized Gram matrix, forming a probability distribution such that $\sum \tilde{\lambda}_{j} = 1$.  This unified definition  allows us to rigorously measure whether SFT preserves, compresses, or displaces the information encoded in the pre-trained features.

\subsection{Optimization Dynamics}\label{sec:optimization}
While the metrics in subsequent sections characterize the SFTs in the representation manifold, the underlying mechanism originates in the optimization landscape. We posit that the intensity of parameter updates reflects the adaptation effort allocated to each layer. Let $\Theta^{(l)}=\{W_{Q},W_{K},W_{V},W_{O}\}^{(l)}$ denote the set of all learnable projection matrices of attention module within the $l$-th Transformer layer, where $W_Q$, $W_K$, $W_V$, and $W_O$ denote the query, key, value, and output 
projection weight matrices of the self-attention module, respectively. To quantify the magnitude of this re-optimization, we define the weights change metric $\Delta\mathcal{W}^{(l)}$ as the aggregated Frobenius distance between the parameter configurations of the SFT model ($\mathcal{M}_{s}$) and the Base model ($\mathcal{M}_{b}$):
\begin{equation}
    \Delta\mathcal{W}^{(l)}=\sqrt{\sum_{W\in\Theta^{(l)}}||W_{s}-W_{b}||_{F}^{2}}
\end{equation}
A high value of $\Delta\mathcal{W}^{(l)}$ indicates that the layer has experienced aggressive parameter modifications induced by the supervised fine-tuning objective. Our central hypothesis is that the layers closest to the output loss function are forced to undergo the most significant structural changes to accommodate the new task constraints. This parameter-level reconstruction acts as the driving force, physically displacing the pre-trained weights and thereby causing the information compression and geometric restructuring observed in the representation space.

\subsection{Information Dynamics}

The  parameter updates described in Section \ref{sec:optimization} inevitably alter the information capacity of the representation space. We employ the matrix-based entropy framework to monitor this SFT, specifically testing the information bottleneck hypothesis: that SFT forces the model to compress generic pre-training features (old information) to accommodate task-specific constraints (new information).

\paragraph{Prompt Entropy.} To quantify the intra-sequence information density and determine whether SFT compresses fine-grained token details, we compute the average entropy over all $N$ samples. For the $i$-th sample representation $H_{i}^{(l)}$, the prompt entropy is:
\begin{equation}
    S_{p}^{(l)}(i) = S_{\alpha}(H_{i}^{(l)}(H_{i}^{(l)})^{\top})
\end{equation}
A decrease in $S_{p}^{(l)}$ ($\Delta < 0$) suggests that SFT induces token compression, filtering out redundant pre-training noise to focus on task-relevant signals.

\paragraph{Dataset Entropy.} To assess the inter-sample diversity and verify if the strong supervision signal causes mode collapse, we compute the entropy of the dataset-level matrix $\overline{H}^{(l)}$:
\begin{equation}
    \mathcal{S}_{d}^{(l)} = S_{\alpha}(\overline{H}^{(l)}(\overline{H}^{(l)})^{\top})
\end{equation}
A negative SFT $\Delta\mathcal{S}_{d}^{(l)} < 0$ indicates that SFT pulls sample representations closer to form tighter, task-specific manifolds.

\paragraph{Effective Rank and Deficiency.} To gauge the true dimensionality of the representation space beyond simple rank constraints, we utilize the effective rank. Let $\sigma_{1} \ge \sigma_{2} \ge ... \ge \sigma_{r} > 0$ be the singular values of the dataset matrix $\overline{H}^{(l)}$. We define the normalized singular value distribution as $p_{j} = \sigma_{j}^{2} / \sum_{k} \sigma_{k}^{2}$. The effective rank is:
\begin{equation}
    \text{EffRank}(\overline{H}^{(l)}) = \exp\left( -\sum_{j=1}^{r} p_{j} \log p_{j} \right)
\end{equation}
We complement this with the algebraic rank deficiency, defined as $Def = \min(N, D) - |r|$. A reduction in effective rank or an increase in deficiency signifies that the model is utilizing a smaller, more efficient subspace to encode the task, confirming the displacement of the broad pre-training basis.

Beyond spectral properties, we examine the raw encoding efficiency through Sparsity, which measures the fraction of inactive neurons ($|z| < \epsilon$) averaged over all dimensions. Higher sparsity indicates  SFT is performing explicit feature selection, pruning irrelevant pre-training neurons to dedicate capacity solely to the target instruction logic.


\subsection{Geometric Restructuring}

Finally, we analyze the external manifestation of these internal SFTs.  First, we analyze how the model navigates the semantic space across a sequence. Let $v_{t}=h_{t}^{(l)}-h_{t-1}^{(l)}$ be the difference vector between consecutive tokens.  {The  curvature quantifies the smoothness of the reasoning path:}
\begin{equation}
\mathcal{C}_i^{(l)}=\frac{1}{(T-2)\pi}\sum_{t=1}^{T-2}\arccos\left(\frac{v_{t}^{\top}v_{t+1}}{||v_{t}||\cdot||v_{t+1}||}\right)
\end{equation}
where $\pi$ is normalization factor, $\mathcal{C}^{(l)}=\frac{1}{N}\sum_{i=1}^{N}\mathcal{C}_i^{(l)}$. A reduction in curvature ($\Delta\mathcal{C} < 0$)  enables the model to maintain more coherent long-term dependencies by simplifying the route through the representation space.

To determine if SFT merely rotates the representation space (isomorphic transformation) or fundamentally restructures it, we employ Centered Kernel Alignment (CKA). Given $K_b = \overline{H}_b^{(l)} (\overline{H}_b^{(l)})^\top$ and $K_s = \overline{H}_s^{(l)} (\overline{H}_s^{(l)})^\top$ be the dataset-level Gram matrices. We first introduce the centering matrix $J = I_N - \frac{1}{N}\mathbf{1}\mathbf{1}^\top$ to remove the mean component. The centered Gram matrices are $\tilde{K}_b = J K_b J$ and $\tilde{K}_s = J K_s J$. The CKA similarity is computed as:
\begin{equation}
    \mathcal{A}_{CKA}^{(l)} = \frac{\langle \tilde{K}_b, \tilde{K}_s \rangle_F}{\| \tilde{K}_b \|_F \| \tilde{K}_s \|_F} = \frac{\mathrm{tr}(\tilde{K}_b \tilde{K}_s)}{\sqrt{\mathrm{tr}(\tilde{K}_b^2) \cdot \mathrm{tr}(\tilde{K}_s^2)}}
\end{equation}
A value of $\mathcal{A}_{CKA}^{(l)} \ll 1$ signals  that the original manifold structure has been substantially restructured to accommodate the new task-specific 
representations.

While CKA captures global structural similarity, we measure explicit directional reorientations using cosine similarity and mean shift. Let $\mu^{(l)}$ denote the centroid of the layer representations.
\begin{equation}
\begin{aligned}
\mathcal{S}_{cos}^{(l)} &= \frac{1}{N} \sum_{i=1}^{N} 
    \frac{ (\overline{h}_{b,i}^{(l)})^{\top} \overline{h}_{s,i}^{(l)} }
         { \|\overline{h}_{b,i}^{(l)}\|_{2} \|\overline{h}_{s,i}^{(l)}\|_{2} }, \\
\mathcal{D}_{SFT}^{(l)} &= \|\mu_{s}^{(l)} - \mu_{b}^{(l)}\|_{2}
\end{aligned}
\end{equation}
A sharp decay in $\mathcal{S}_{cos}^{(l)}$ coupled with a spike in $\mathcal{D}_{SFT}^{(l)}$ provides the  coordinate-level evidence that the representation has been physically transported to a new region of the vector space, driven by the optimization dynamics described in Section \ref{sec:optimization}.
\begin{figure*}[!t]
    \centering
    \includegraphics[width=0.32\linewidth]{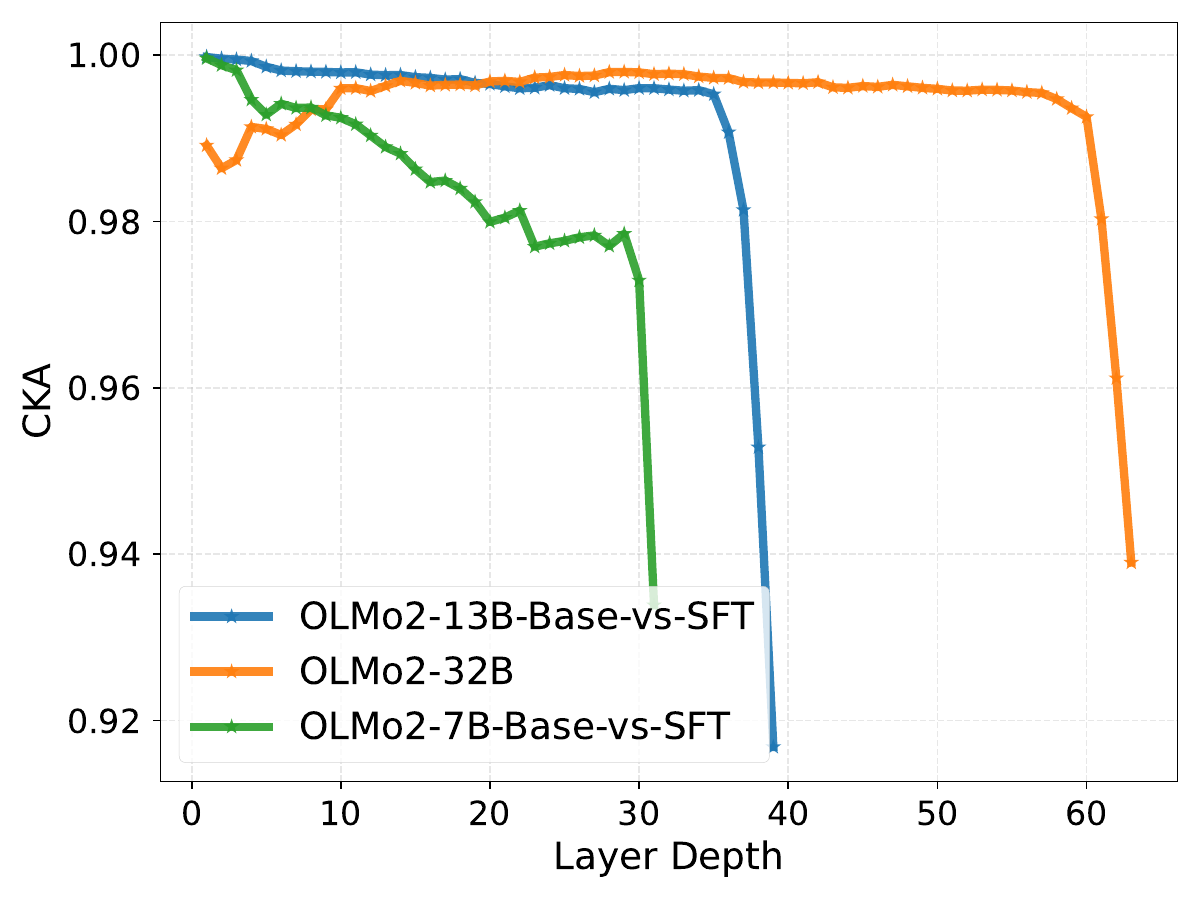}
    \includegraphics[width=0.32\linewidth]{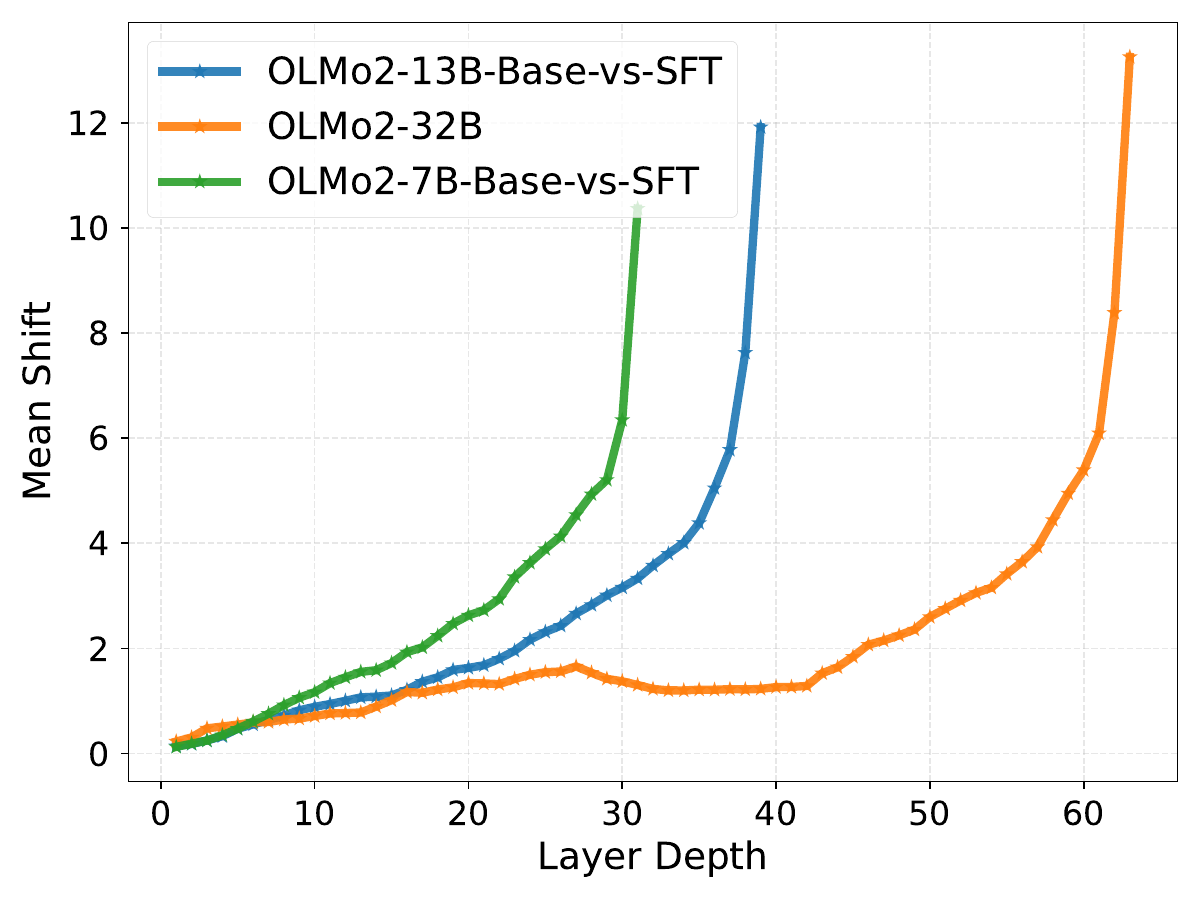}
    \includegraphics[width=0.32\linewidth]{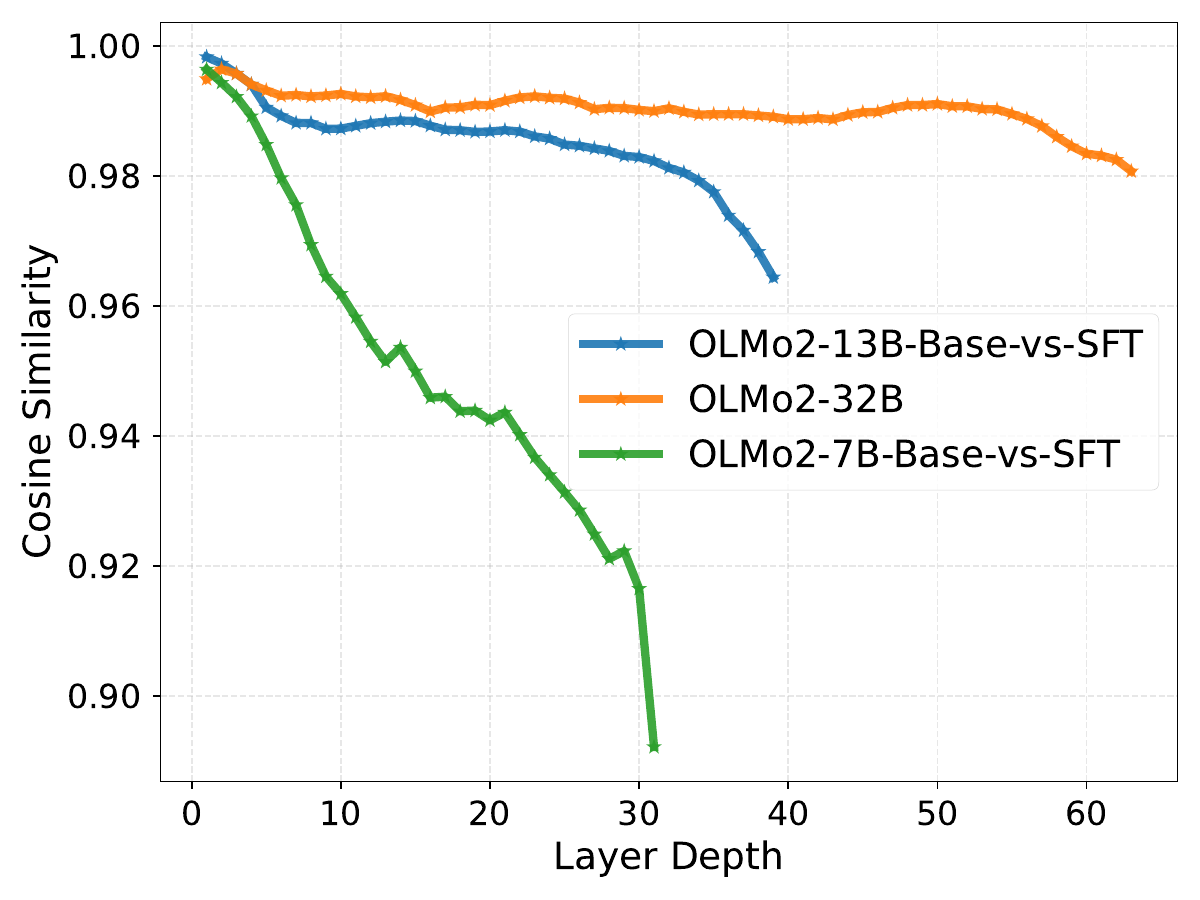}
    \caption{Representational divergence metrics (CKA, Cosine Similarity and Mean Shift) across  layers.}
    \label{fig:similarity}
\end{figure*}
\subsection{Evaluation Protocol}

To ensure rigorous reproducibility and properly quantify the evolutionary gap between the Base and SFT models, we categorize our analysis into three computation modes based on the aggregation scope of the representations. Let $f_{k}$ denote a specific metric function defined in the previous sections. We strictly distinguish between local token dynamics and global manifold properties.

\begin{itemize}
    \item \textbf{Sample-Level Difference } For metrics quantifying local properties (e.g., prompt entropy, curvature, cosine similarity), we compute the change per sample and report the average. This captures the mean intra-sample evolutionary magnitude:
    \begin{equation}
        \Delta Q_{\textit{s}}^{(l)}=\frac{1}{N}\sum_{i=1}^{N}(f_{k}(H_{s,i}^{(l)})-f_{k}(H_{b,i}^{(l)}))
    \end{equation}
    \item \textbf{Dataset-Level Difference } For global properties dependent on the full distribution (e.g., dataset entropy, effective rank), we apply the metric to the holistic dataset matrix to prevent aggregation artifacts:
    \begin{equation}
        \Delta Q_{\textit{d}}^{(l)}=f_{k}(\overline{H}_{s}^{(l)})-f_{k}(\overline{H}_{b}^{(l)})
    \end{equation}
    \item \textbf{Alignment Score } For interaction metrics that require dual inputs (e.g., CKA), we measure the geometric correspondence directly without subtraction:
    \begin{equation}
        A_{\textit{a}}^{(l)}=f_{k}(\overline{H}_{b}^{(l)},\overline{H}_{s}^{(l)})
    \end{equation}
\end{itemize}


\section{Experiments}
In this section, we conduct a comprehensive analysis to understand the impact of SFT on model representations and leverage these insights to propose a more efficient tuning strategy.

\subsection{Experimental Setup}
\paragraph{Models} 
We focus on model families that provide both pre-trained (Base) and SFT checkpoints. We utilize Mistral-7B and the OLMo2 series (scaling from 1B to 32B), to rigorously analyze the impact of SFT without the confounding factors of complex alignment techniques (e.g., RLHF or DPO).
To further contextualize the model selection in our experimental setup, we provide a  survey of commonly used open-source LLMs and their alignment pipelines, justifying why OLMo2 and Mistral-7B are the most suitable choices for isolating pure SFT dynamics. Details are provided in Appendix~\ref{appendix:modelsurvey}.
We conduct experiments using official checkpoints from the Hugging Face Hub to ensure reproducibility. Specifically, we evaluate {mistralai/Mistral-7B-v0.1} (Base) and its Instruct-v0.1 (SFT) variant. For the OLMo 2 family, we utilize the entire suite including {allenai/OLMo-2-0425-1B}, {allenai/OLMo-2-1124-7B}, {allenai/OLMo-2-1124-13B}, and {allenai/OLMo-2-0325-32B} across both Base and SFT versions.

\paragraph{Datasets}  To ensure a comprehensive evaluation of model capabilities, we employ a diverse suite of benchmarks: Massive Multitask Language Understanding (MMLU, ~\cite{hendrycks2021measuring}), Grade School Math 8K (GSM8K,~\cite{cobbe2021trainingverifierssolvemath}), WikiText Language Modeling Dataset (WikiText, ~\cite{merity2017pointer}), Instruction Following Evaluation (IFEval,~\cite{zhou2023instructionfollowingevaluationlargelanguage}), HumanEval~\citep{chen2021evaluatinglargelanguagemodels}, Multi-Turn Benchmark (MT-Bench,~\citealt{zheng2023judging}), and ToxiGen~\citep{hartvigsen-etal-2022-toxigen}. We categorize these into two analysis granularities. For token-level analysis (MMLU, GSM8K, WikiText, and IFEval), where representations are extracted for every token in the sequence, we sample 100 instances from the test sets. Conversely, for pooled-level analysis (including all datasets), we adopt a strategy where each input corresponds to a single vector, using a sample size of 1,000 instances from the respective test sets.

\begin{figure*}[!t]
    \centering
    \includegraphics[width=0.32\linewidth]{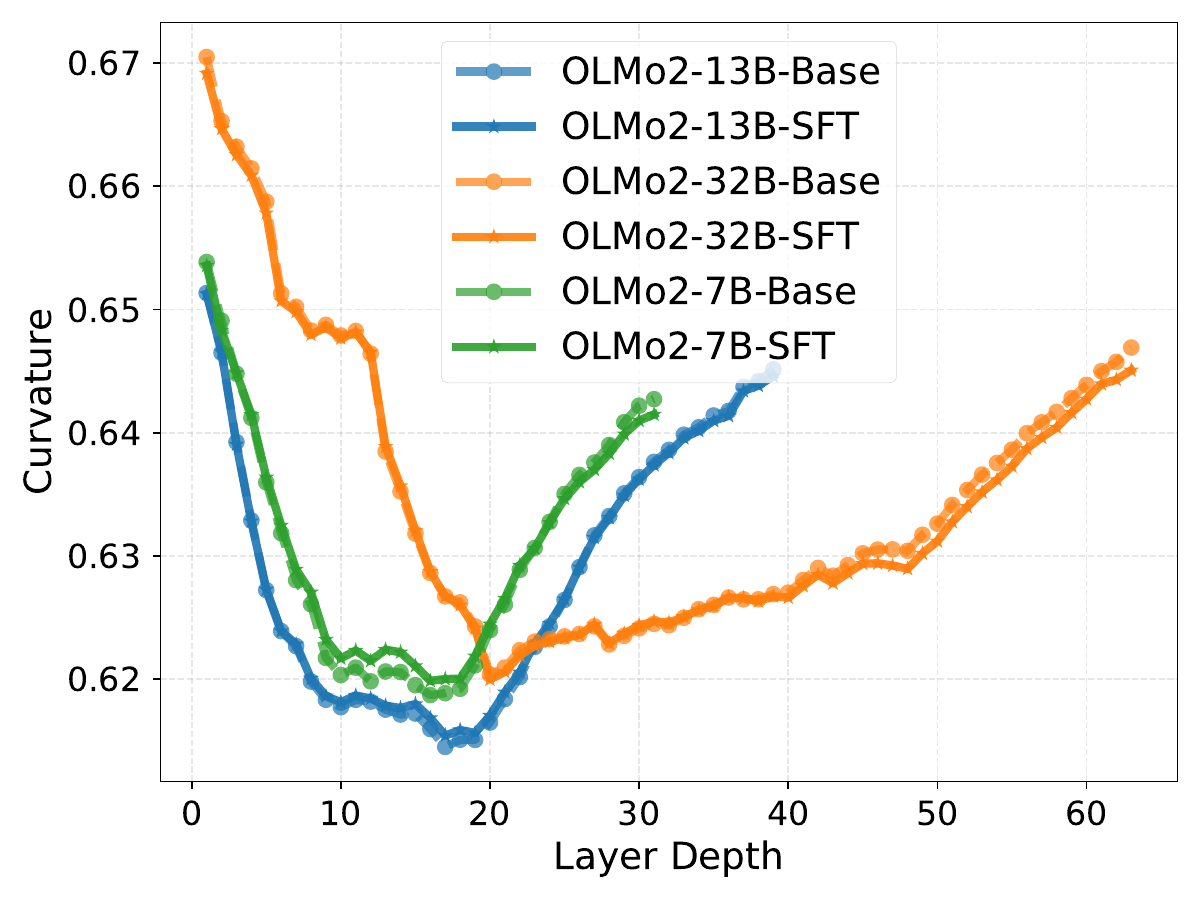}
    \includegraphics[width=0.32\linewidth]{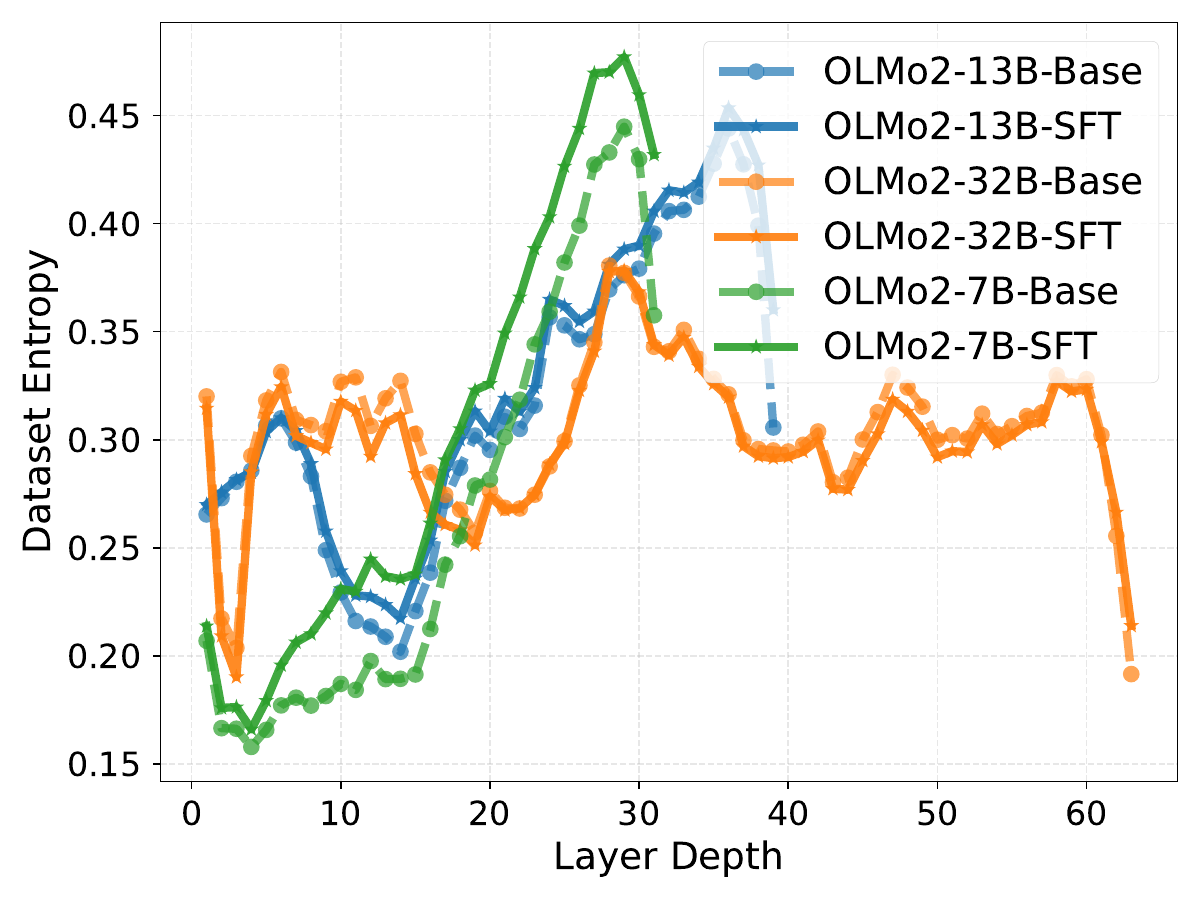}
    \includegraphics[width=0.32\linewidth]{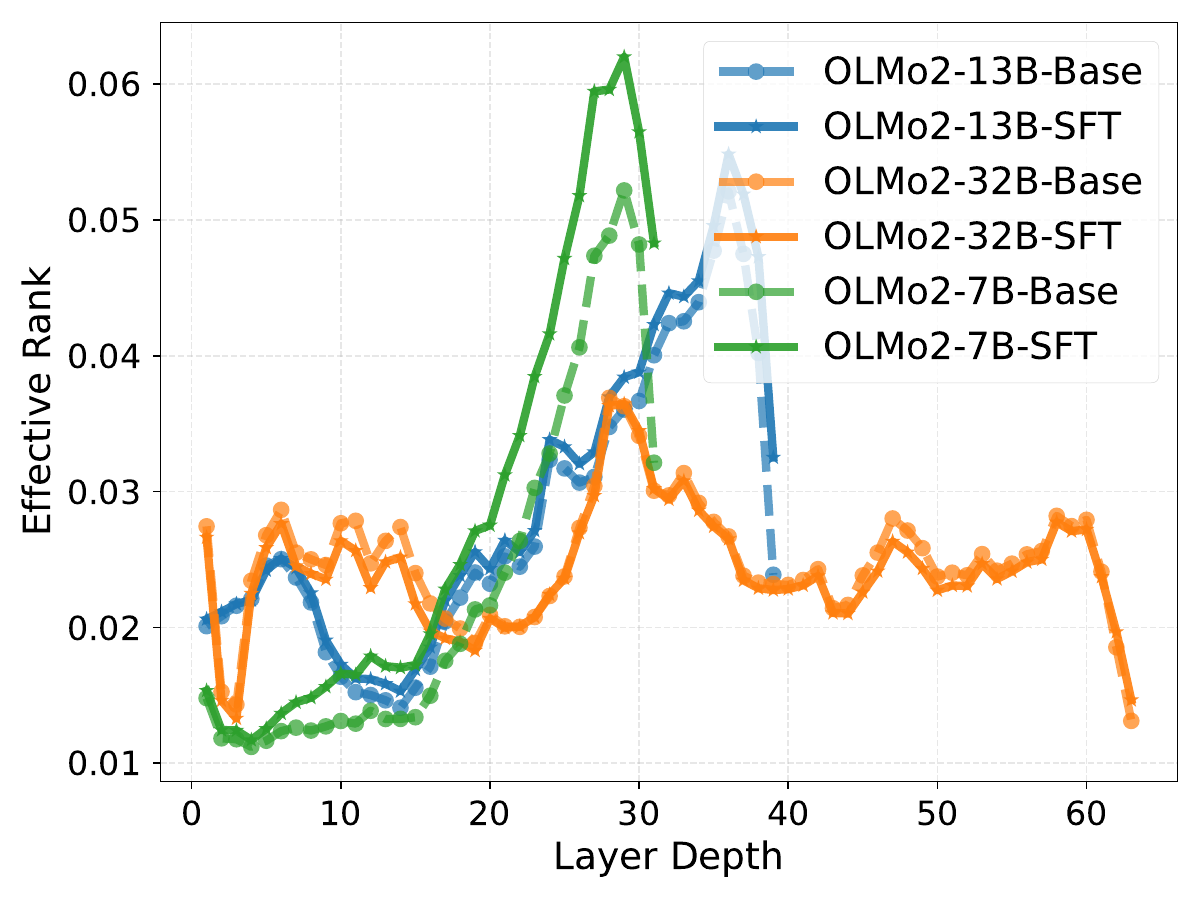}
    \includegraphics[width=0.32\linewidth]{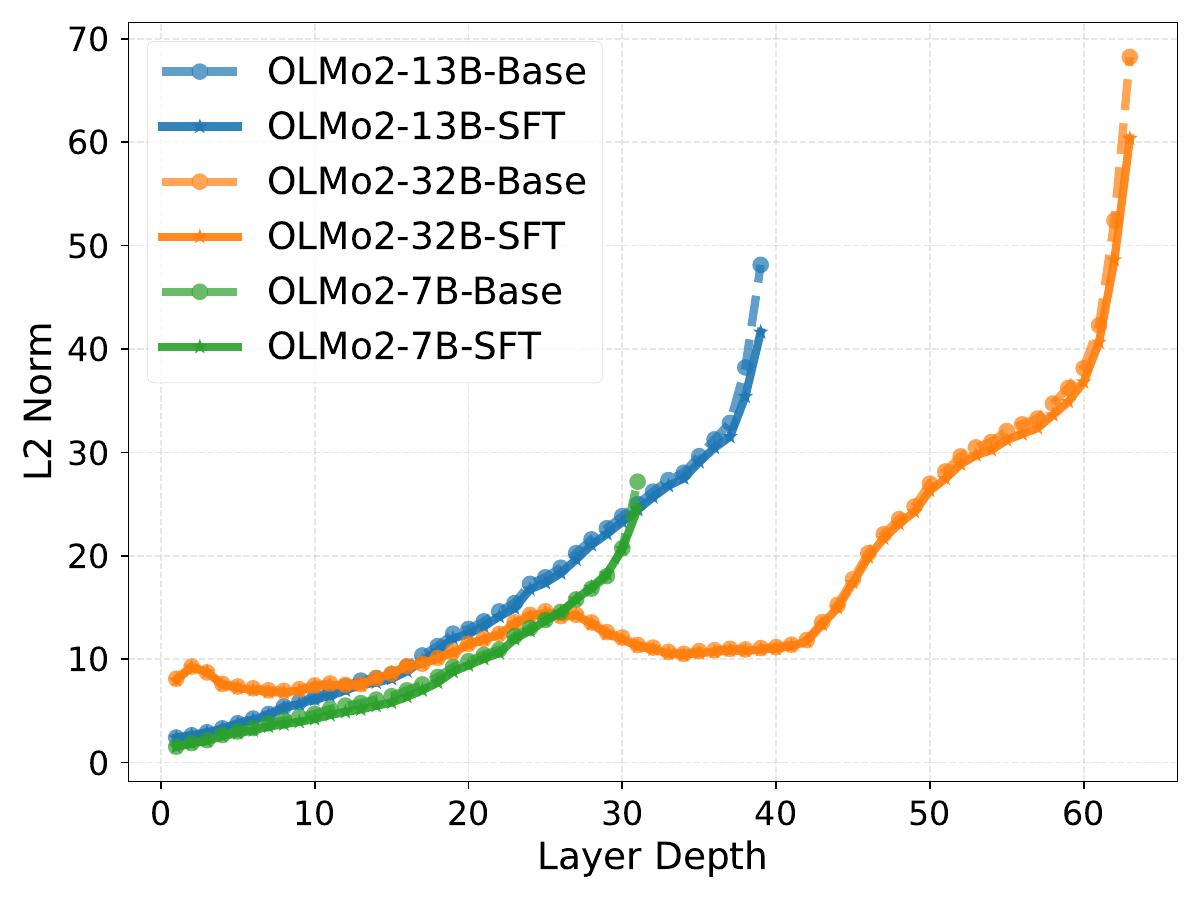}
    \includegraphics[width=0.32\linewidth]{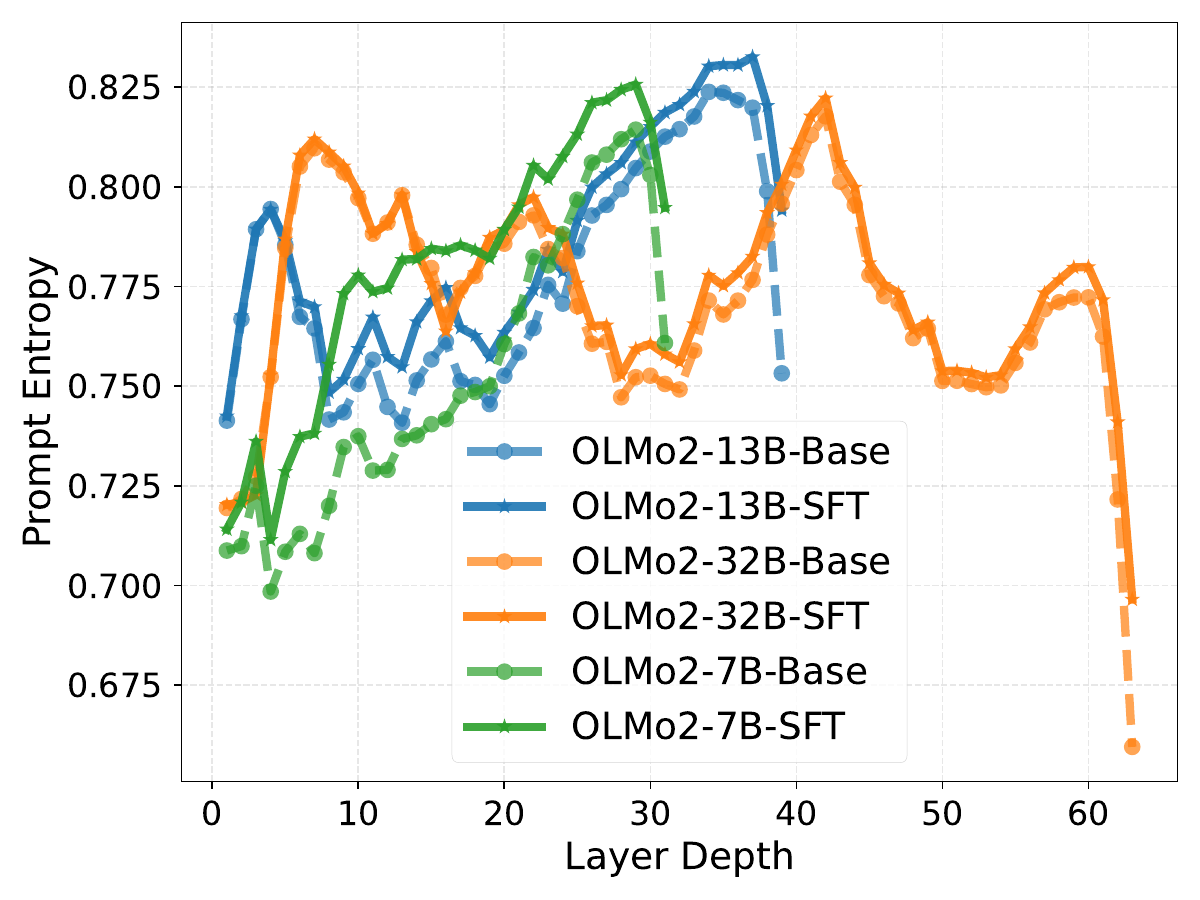}
    \includegraphics[width=0.32\linewidth]{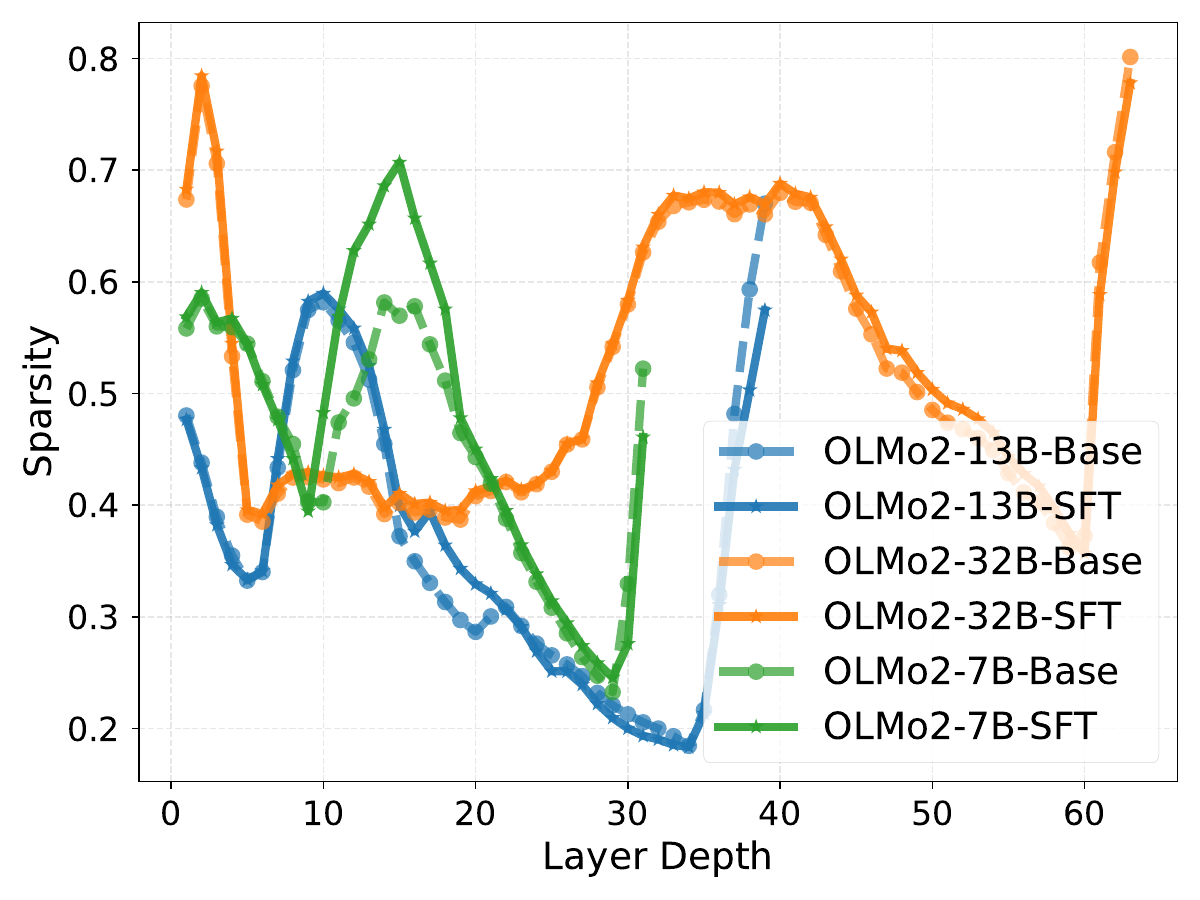}
    \includegraphics[width=0.32\linewidth]{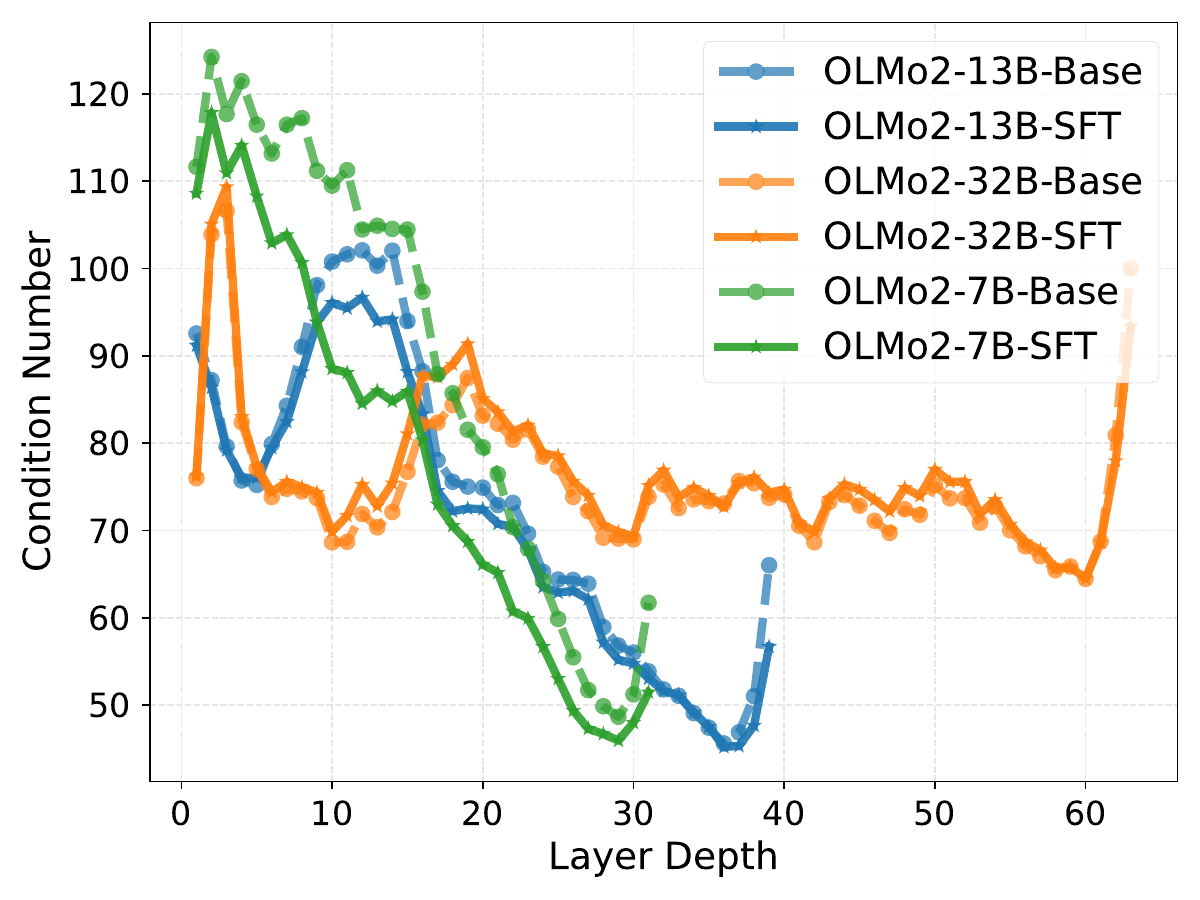}
    \includegraphics[width=0.32\linewidth]{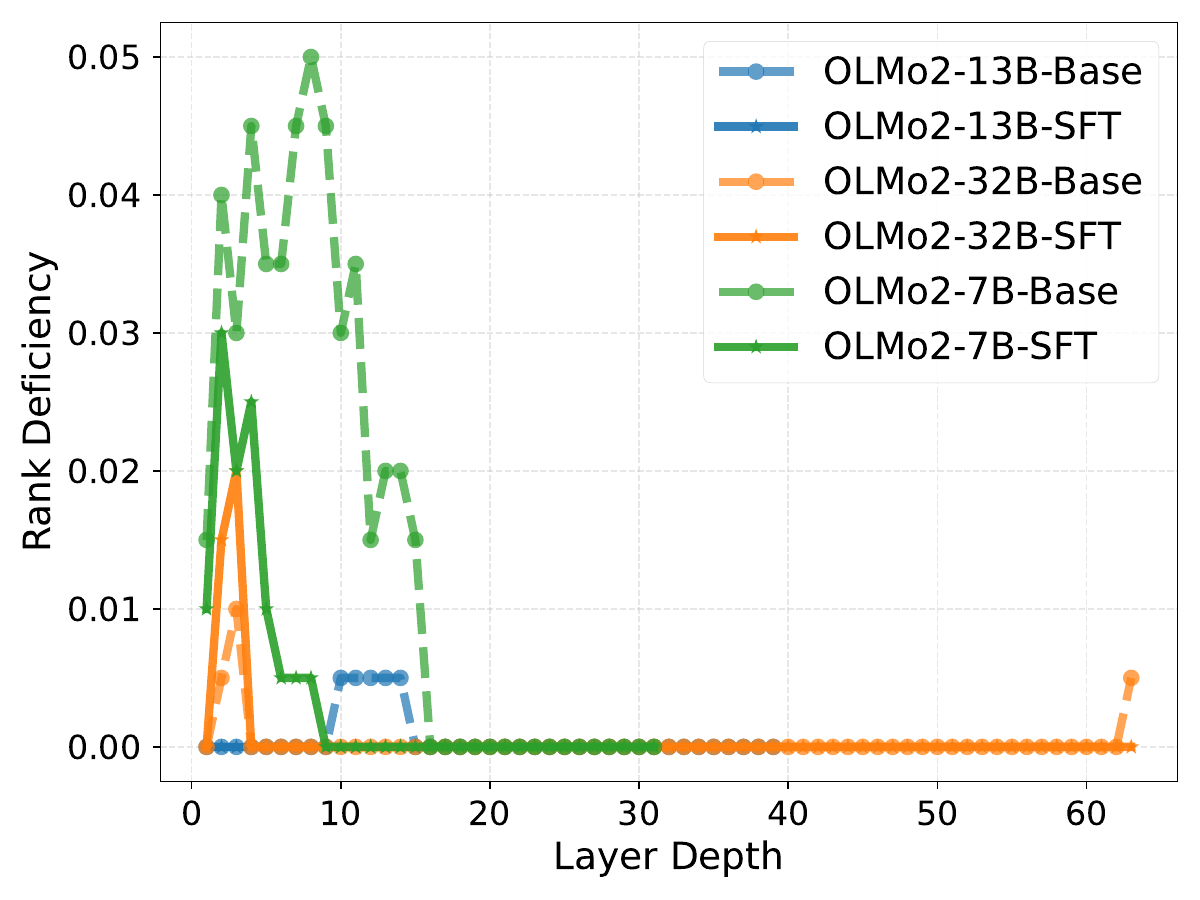}
    \includegraphics[width=0.32\linewidth]{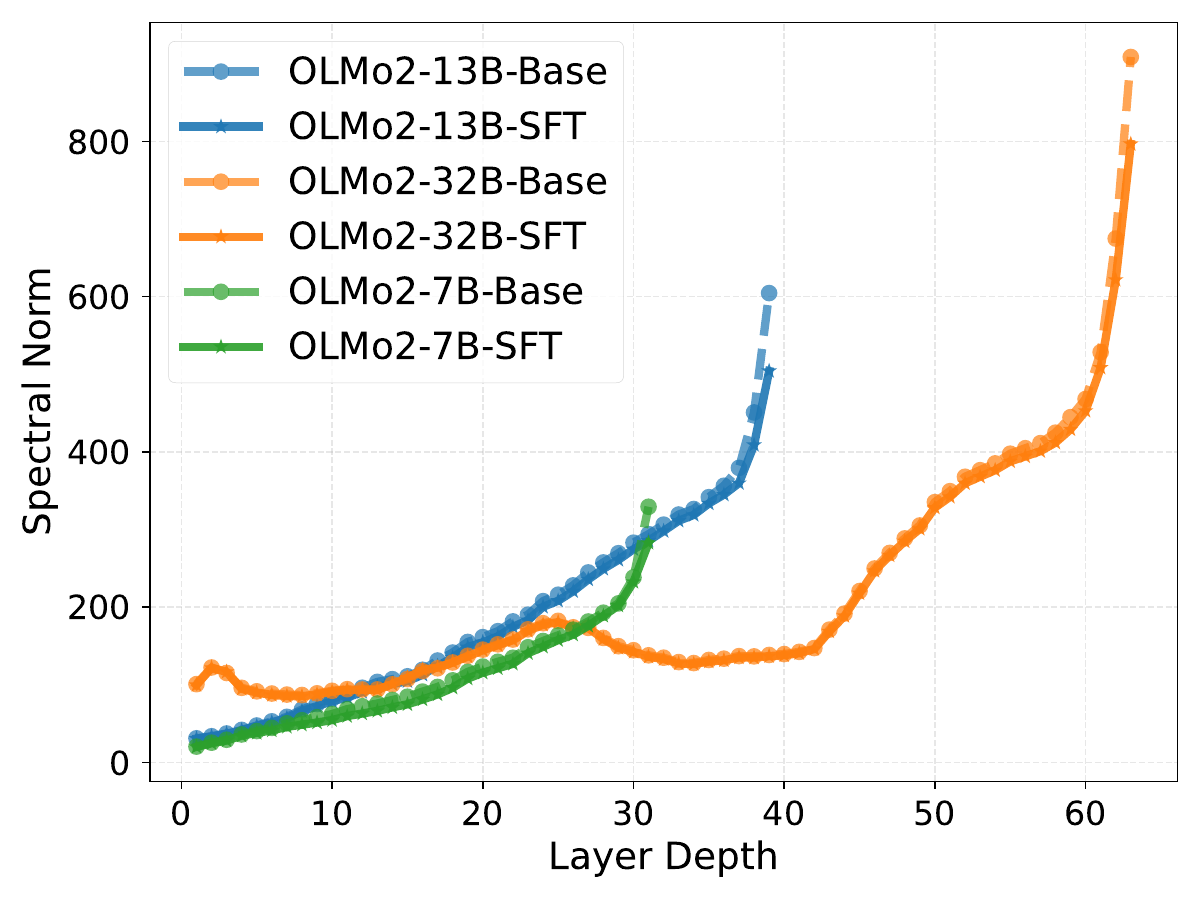}
    \caption{Intrinsic representation metrics across layers for Base and SFT models.}
    \label{fig:curvature}
\end{figure*}

\paragraph{Implementation Details.}
Given the high dimensionality of representations (e.g., $L=40, T=256, D=4096$), storing full representation tensors is computationally infeasible. We thus employ streaming computation to process representations on-the-fly, minimizing memory overhead. To ensure deterministic evaluation, particularly for invariance metrics involving augmentations, we fix global random seeds and strictly enforce sample processing order. Furthermore, for entropy-based metrics, to ensure comparability across layers with varying dimensions, we report the normalized matrix entropy: $\bar{S}_{\alpha}(K) = S_{\alpha}(K) / \log_2(N)$, where $N$ is the batch size. $\epsilon=0.01$ is set to compute representation sparsity.  The code is publicly available at \url{https://anonymous.4open.science/r/base_sft}.

\subsection{{Layer-wise Dynamics of SFT}}\label{sec:dynamics}
To deconstruct the evolutionary trajectory induced by supervised fine-tuning, we first investigate how SFT alters the internal representations compared to the Base model across different layers.

\paragraph{Interact-Divergence} 
We quantify the divergence between the Base and SFT models using Cosine Similarity, CKA and mean shift. As illustrated in Figure \ref{fig:similarity}, distinct evolutionary phases were observed across model scales. 
Taking OLMo2-32B as an example, the CKA remains stable in the shallow layers (0--56), where the score plateau stays above $0.98$. However, this was followed by a sharp drop in deeper layers (56 to the end), where the score decays significantly, dropping to approximately $0.94$ in the final layer.
This directional shift was corroborated by the mean shift ( and also cosine similarity) metric, which remains negligible ($<1.0$) for the majority of the shallow layers (35 vs. 40 layers) but spikes exponentially in the final 5 layers, reaching a magnitude of over $12.0$.
To verify that these SFTs are induced by supervised fine-tuning rather than random initialization noise, we conducted a robustness check using different random seeds. A \textit{t}-test confirms the statistical significance of these differences ($p < 0.01$).

\paragraph{Independent-Convergence}

While the previous analysis highlighted the significant divergence between Base and SFT representations, we examine the layer-wise trends of intrinsic metrics (e.g., effective rank) for both Base and SFT models, as shown in Figure \ref{fig:curvature}. In contrast to the interact-divergence observed in Figure \ref{fig:similarity}, the evolution of Base and SFT models remains remarkably synchronized. Specifically, the curves exhibit a consistent three-stage pattern: an initial transition occurs from the embedding layer (Layer 0) to Layer 1, followed by a relatively stable phase across the extensive intermediate blocks. In this middle zone, the geometry reflects a semantic expansion, where effective rank plateaus at its peak (e.g., $\sim 0.05$ for OLMo2-13B) and the Condition Number drops to a noise-resilient trough of $\sim 45$, physically functioning as a high-dimensional, stable substrate for reasoning. Finally, a distinct functional shift re-emerges within the last $\sim 20\%$ layers, manifesting as an information bottleneck. Here, the effective rank collapses to $<0.01$ while Spectral Norm explodes to over 500, forcefully compressing features into a low-rank, high-magnitude state to drive the decisive output distribution.

\subsection{Locating the Task Adaptation}\label{sec:task_adaption}
The SFT observed in Section \ref{sec:dynamics} prompts us to investigate this through probing, weights analysis, and layer swapping.

\begin{figure*}[!t]
    \centering
    \includegraphics[width=0.245\linewidth]{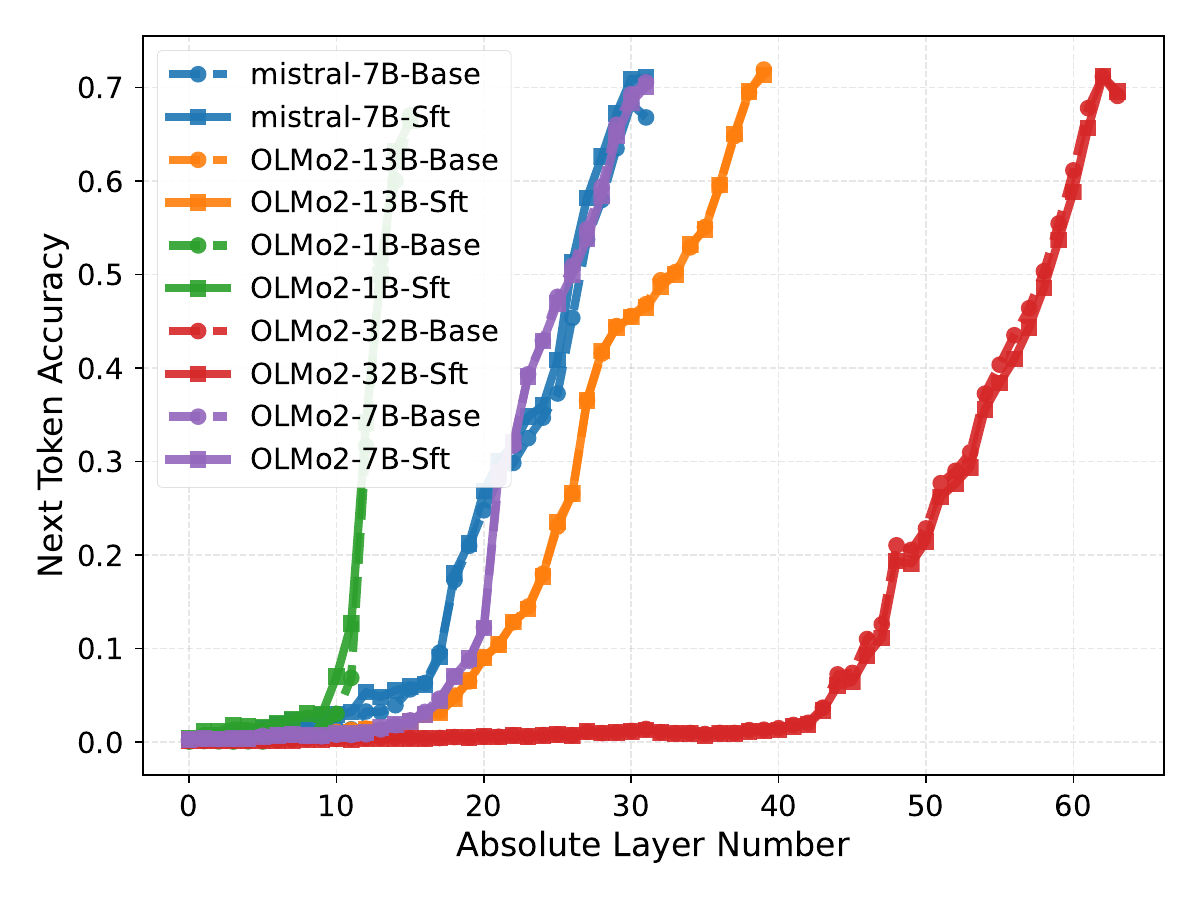}
    \includegraphics[width=0.245\linewidth]{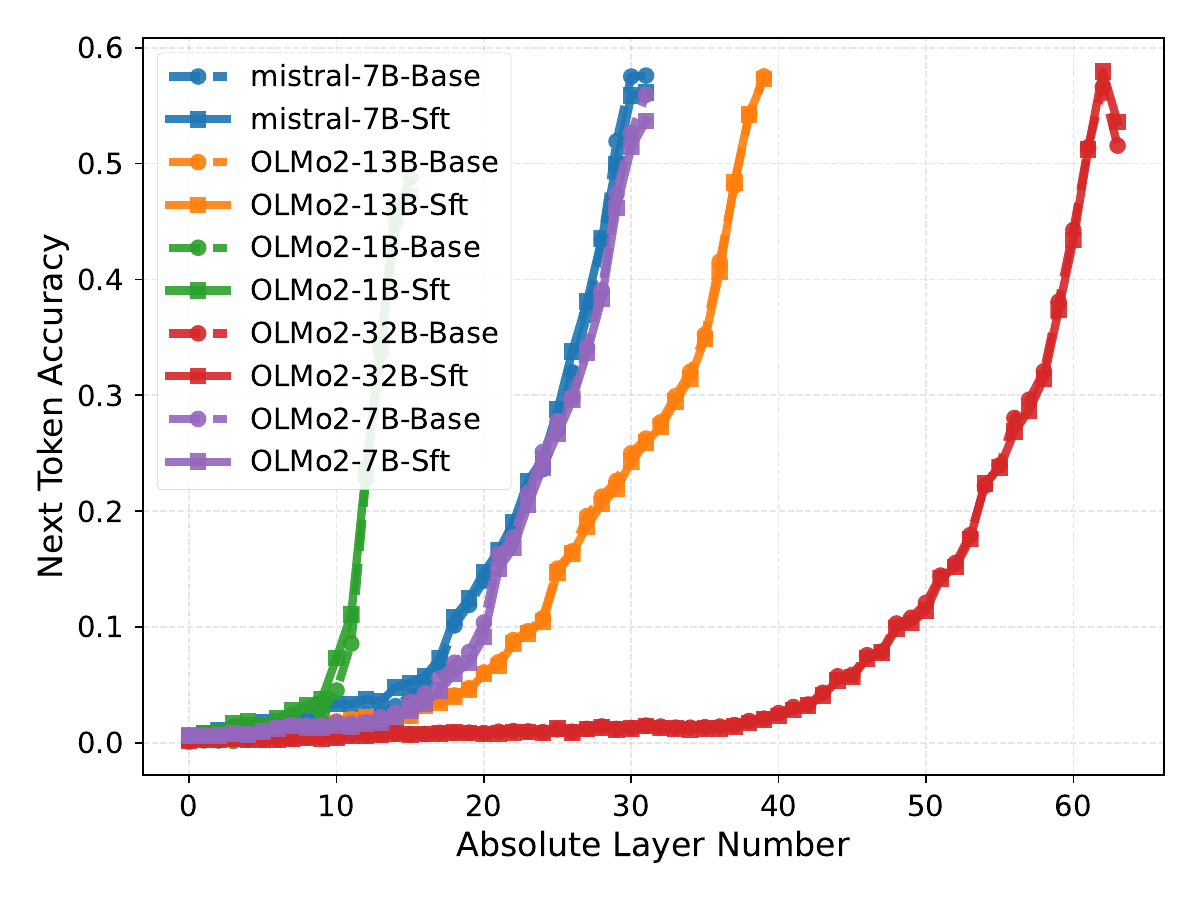}
    \includegraphics[width=0.24\linewidth]{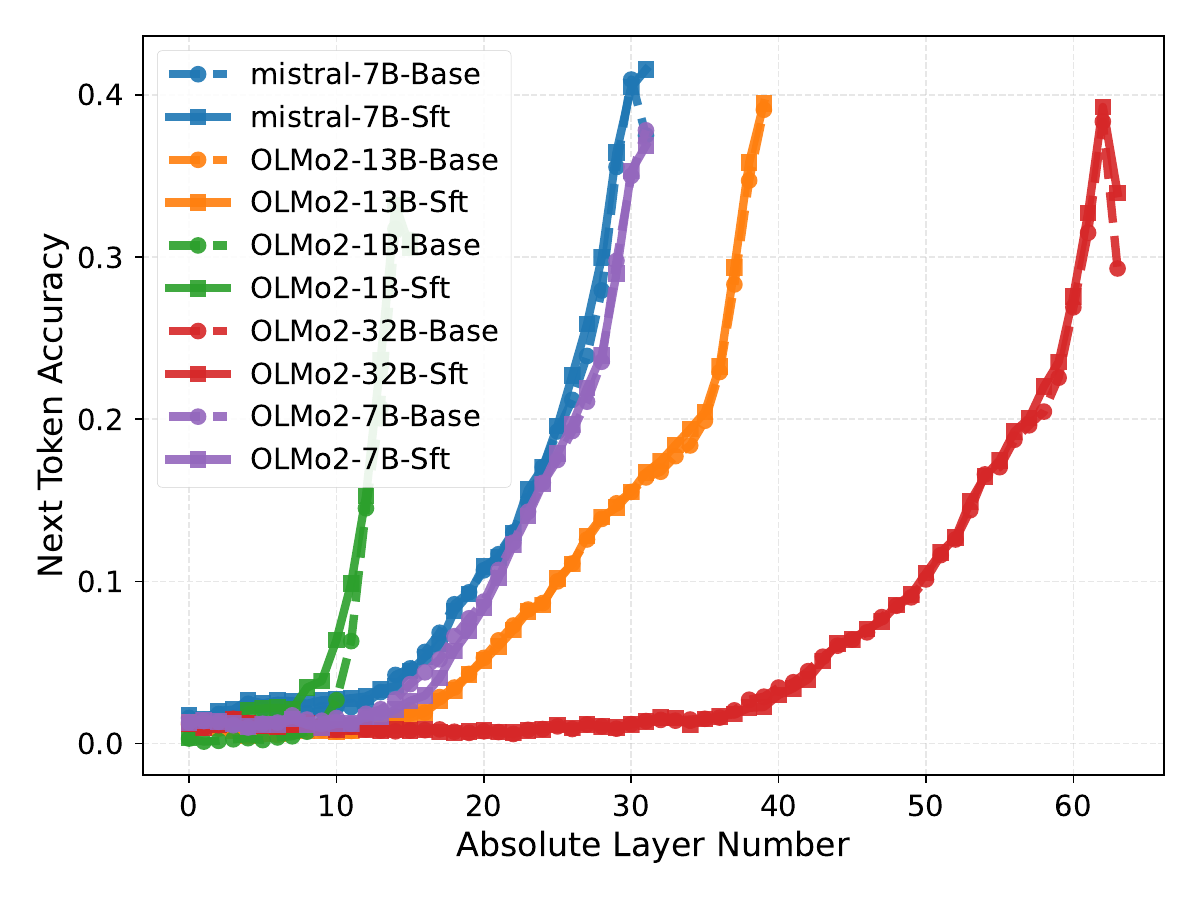}
    \includegraphics[width=0.245\linewidth]{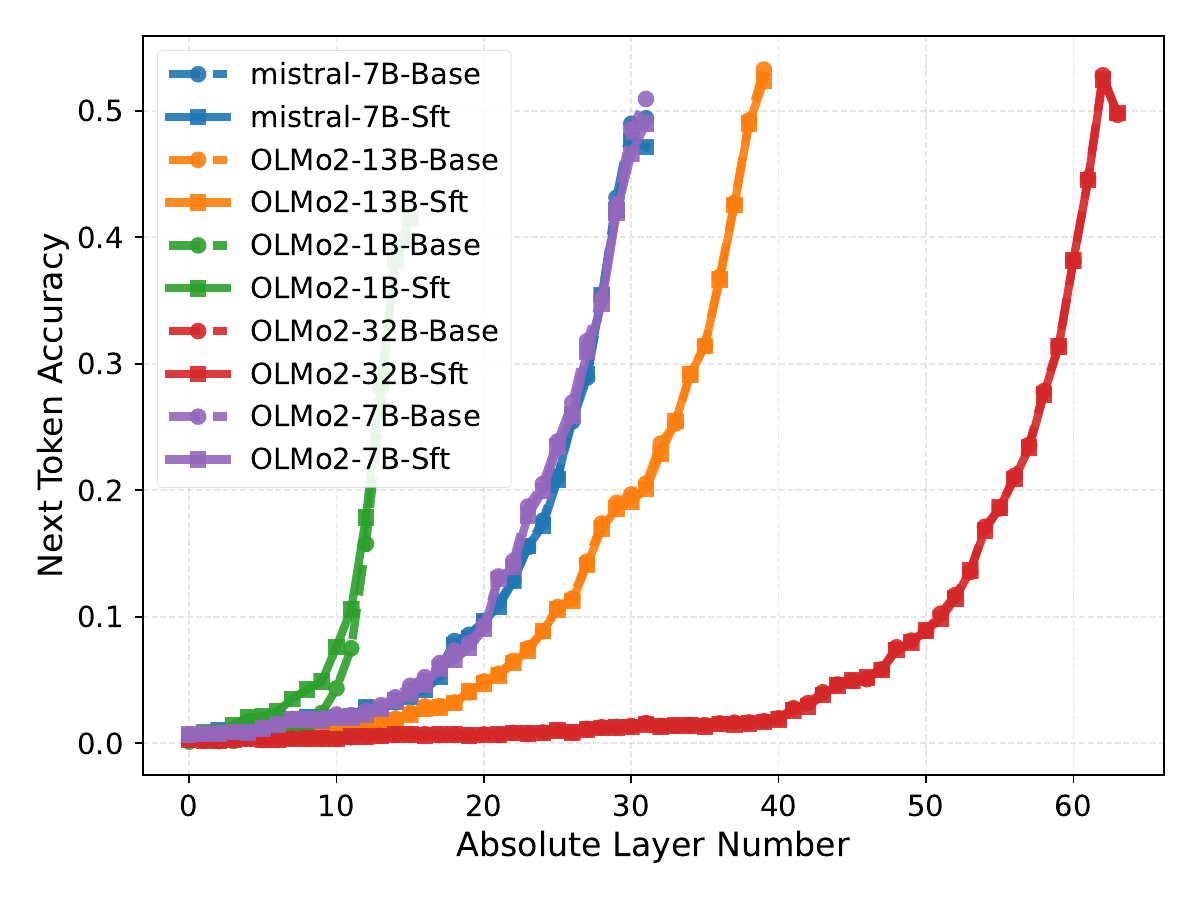}
    \includegraphics[width=0.245\linewidth]{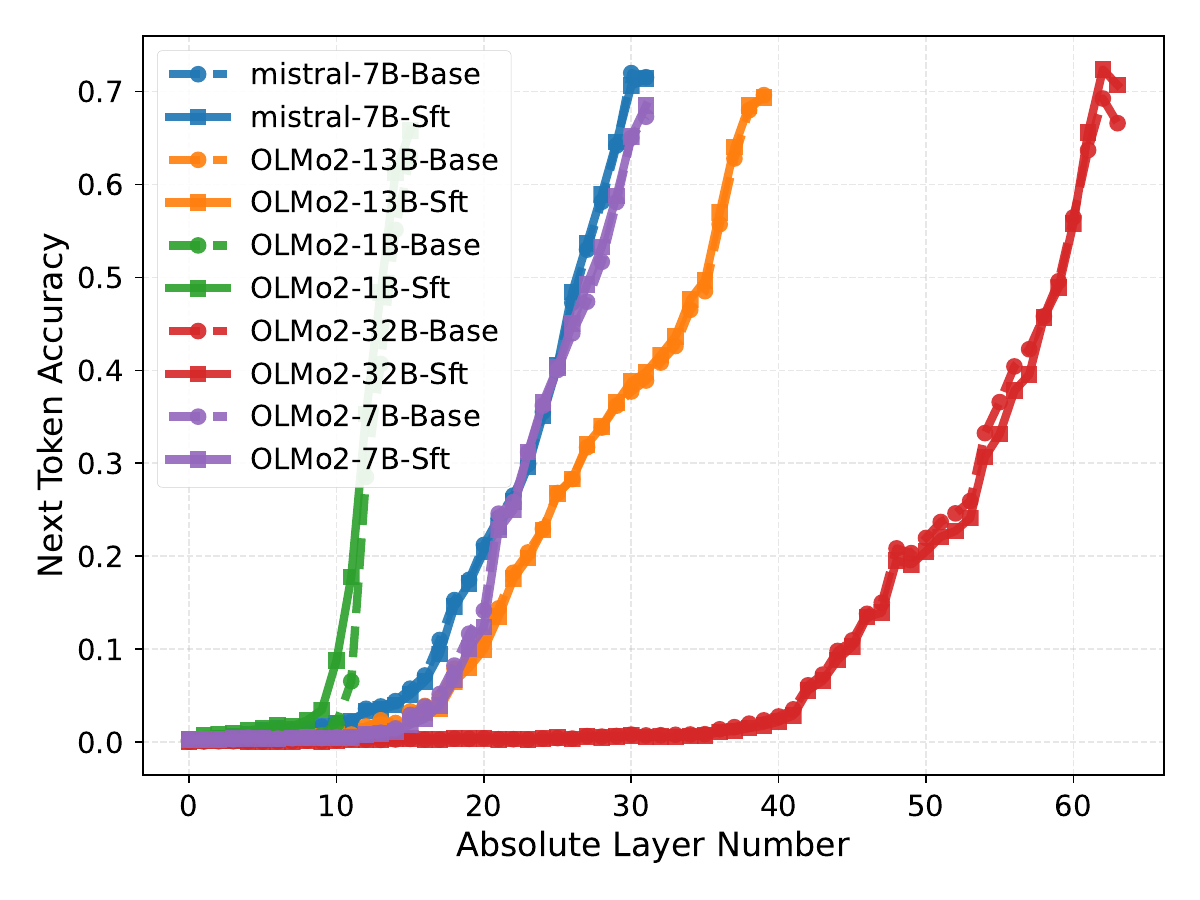}
    \includegraphics[width=0.24\linewidth]{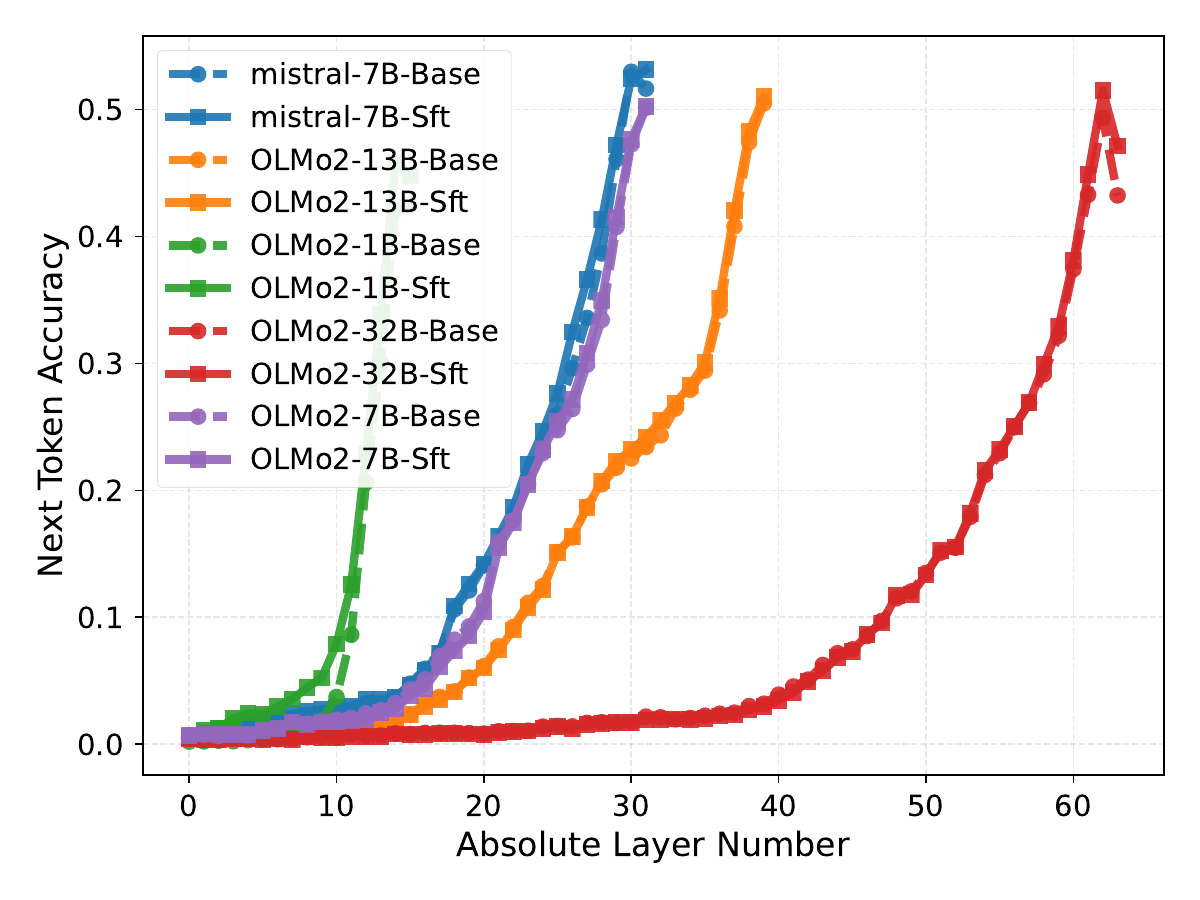}
    \includegraphics[width=0.24\linewidth]{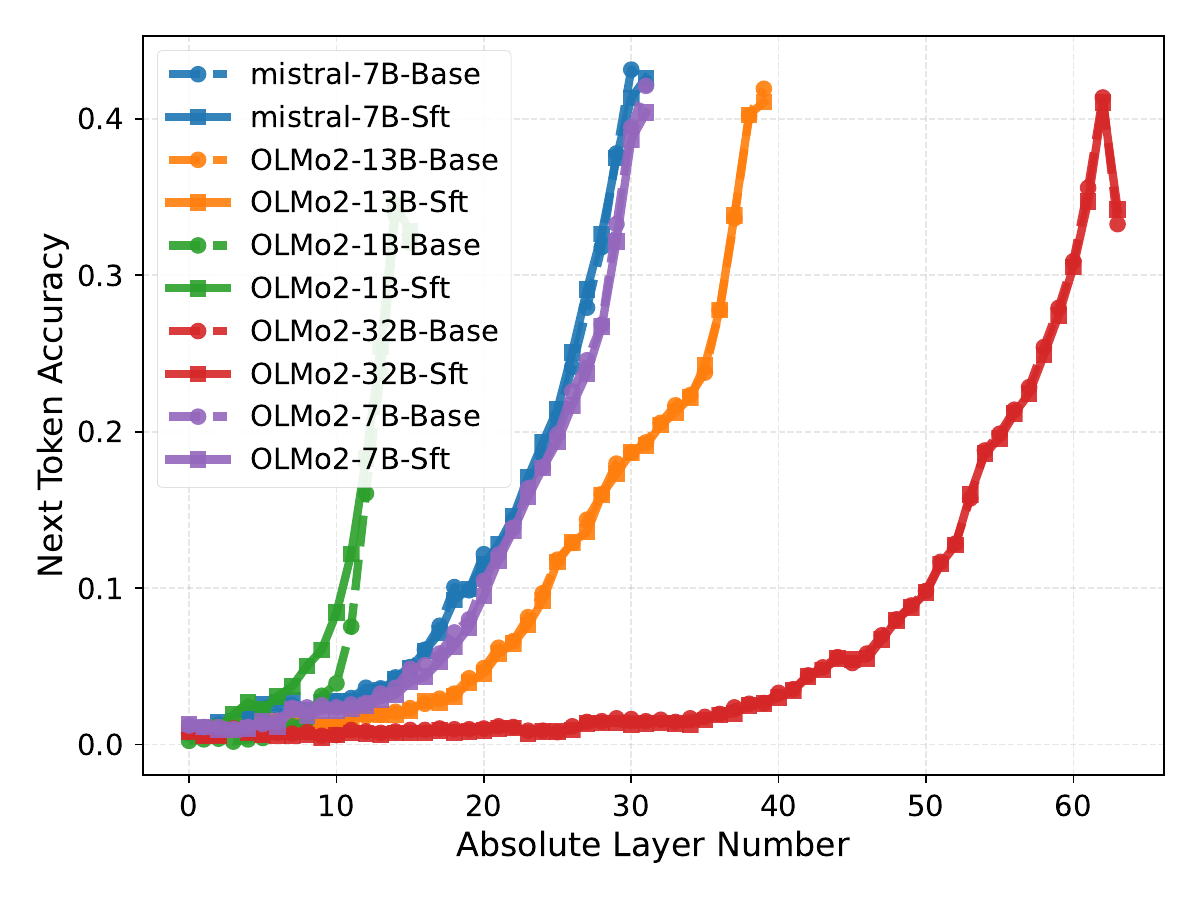}
    \caption{Next-token prediction accuracy of layer-wise probing, from left to right is GSM8K, MMLU, IFEval, WikiText,  HumanEval, MT-Bench, and ToxiGen dataset.}
    \label{fig:probing}
\end{figure*}

\paragraph{Layer-wise Probing}
We hypothesize that the significant representational SFTs in later layers correspond to the emergence of task adaptation capabilities. To verify this, we perform a layer-wise probing experiment where we use the output of each intermediate layer to directly predict the next token.
Figure \ref{fig:probing} displayed the next token prediction accuracy across layers. The results reveal a striking ``dormancy-to-emergence'' pattern, which is particularly pronounced in larger scales.
Taking the {OLMo2-32B} model as a prime example, the probing accuracy remains negligible (below $0.05$) for the vast majority of its depth (the first {50 layers}). However, a sharp phase transition occured in the final block: within the last 14 layers, the accuracy exhibits a vertical ascent, surging from near-zero to over {0.60} on the MMLU dataset. Although a slight decrease is observed in the final layer (from 0.60 to 0.52), we defer analysis of this phenomenon and focus on the transition from mid-to-late layers.

\begin{figure}
    \centering
    \includegraphics[width=\linewidth]{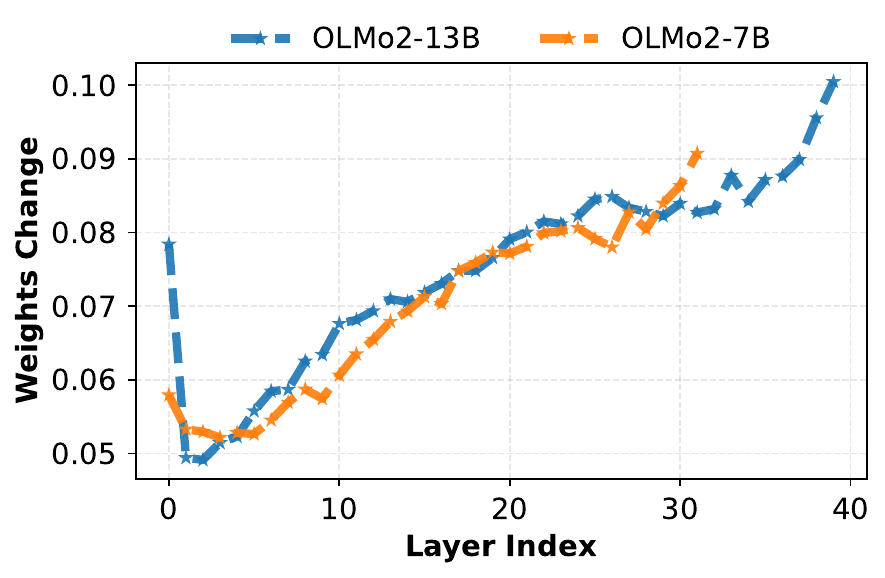}
    \caption{Layer-wise magnitude of weight updates ($L_2$ norm).}
    \label{fig:parameter_change}
\end{figure}
\begin{figure}
    \centering
    \includegraphics[width=0.49\linewidth]{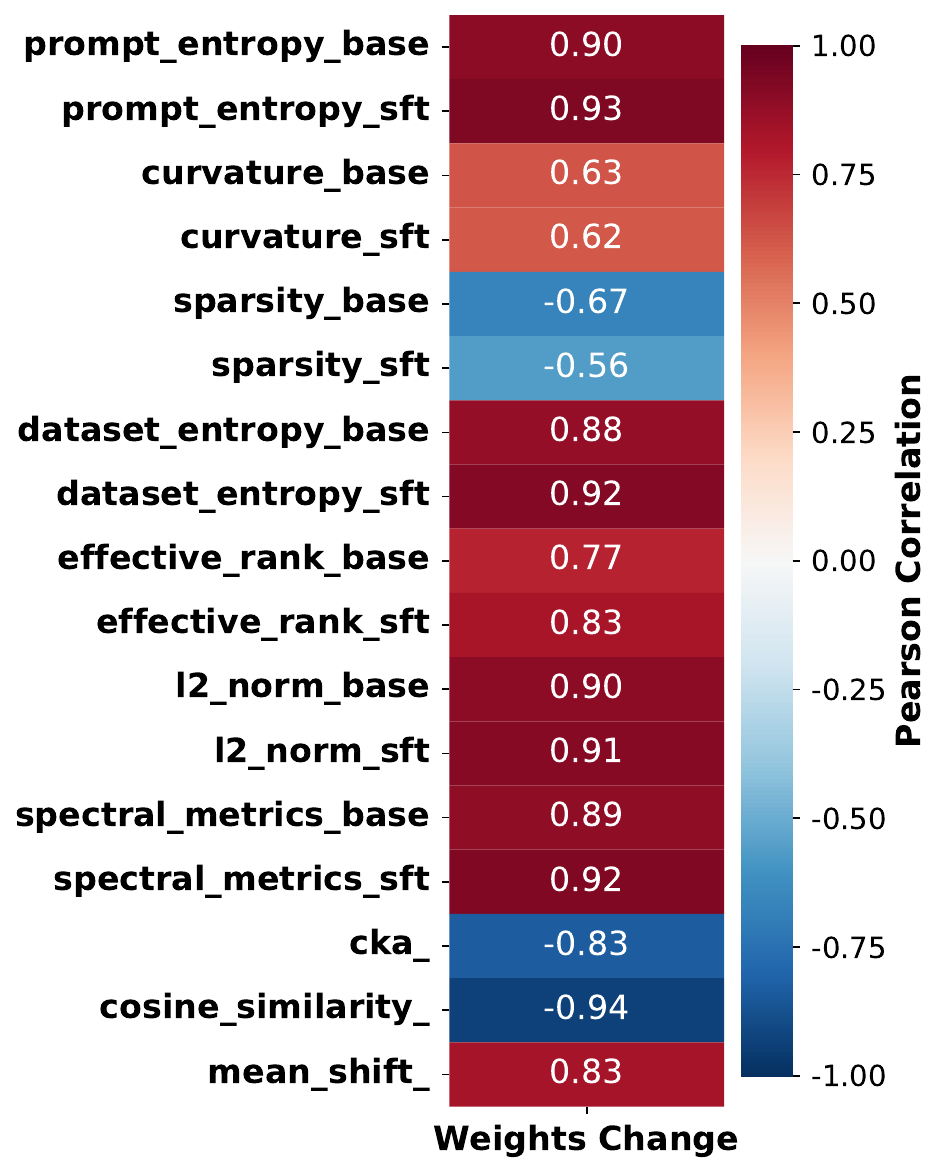}
    \includegraphics[width=0.49\linewidth]{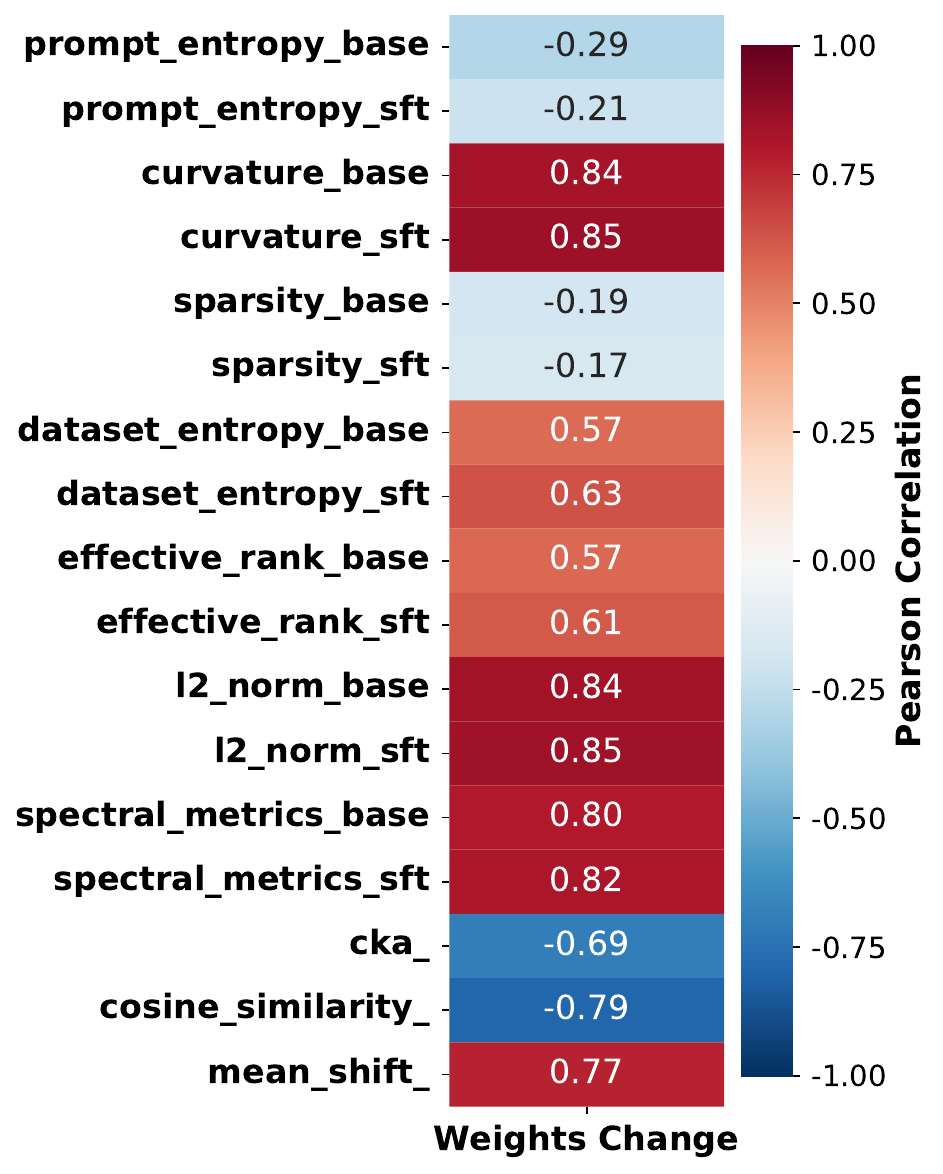}
    \caption{Correlation between layer-wise magnitude of weight updates and  representation  metrics, OLMo2-7B (left), OLMo2-13B (right).}
    \label{fig:correlation}
\end{figure}
\paragraph{Layer-wise Weight Change}
To physically ground the representational shift observed in Section \ref{sec:dynamics}, we revisit the optimization dynamics. We fine-tuned the models on a distinct downstream train dataset and recorded the magnitude of parameter updates ($L_2$ norm of weight changes $\Delta\mathcal{W}^{(l)}$) for each layer.
As shown in Figure \ref{fig:parameter_change}, the update intensity was non-uniform. Taking the {OLMo2-13B} model as an example, the weight change exhibits a J-shaped trajectory: minimal (${\sim}0.05$) in the early layers, it climbs monotonically after the midpoint, reaching a peak of over 0.10 in the final layer. 

The correlation heatmap in Figure \ref{fig:correlation} provided statistical confirmation. Taking OLMo2-13B as an example, we observed a strong negative correlation ($r = -0.79$) between weights change and cosine similarity, and a strong positive correlation ($r > 0.8$) with spectral metrics.
{It maybe due to that, during SFT, the supervision signal from the loss function is strongest at the output and attenuates as it back-propagates. Consequently, the ``knowledge'' required for the new task is preferentially encoded in the late layers through aggressive weight updates, while the bottom layers, shielded by gradient attenuation, implicitly act as a frozen feature extractor.}

\begin{figure}
    \centering
    \includegraphics[width=0.45\linewidth]{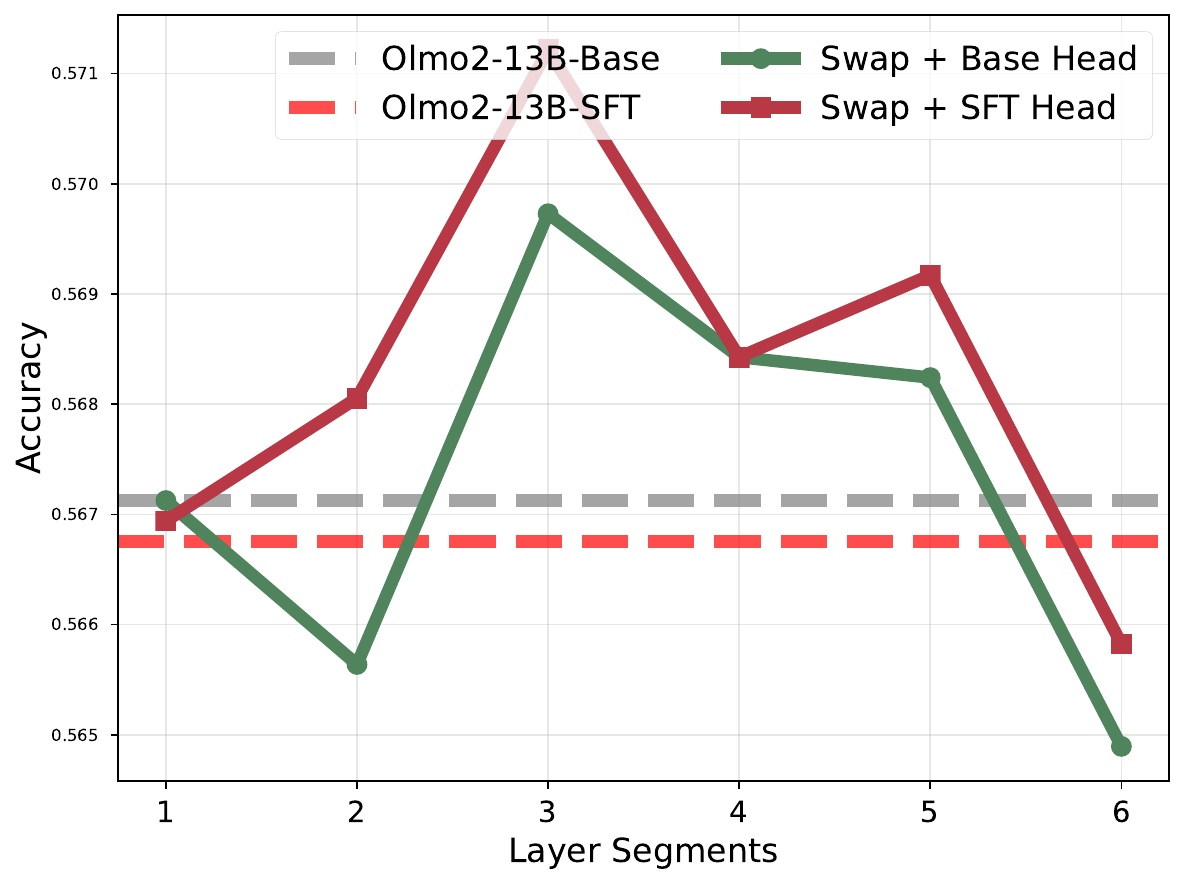}
    \includegraphics[width=0.45\linewidth]{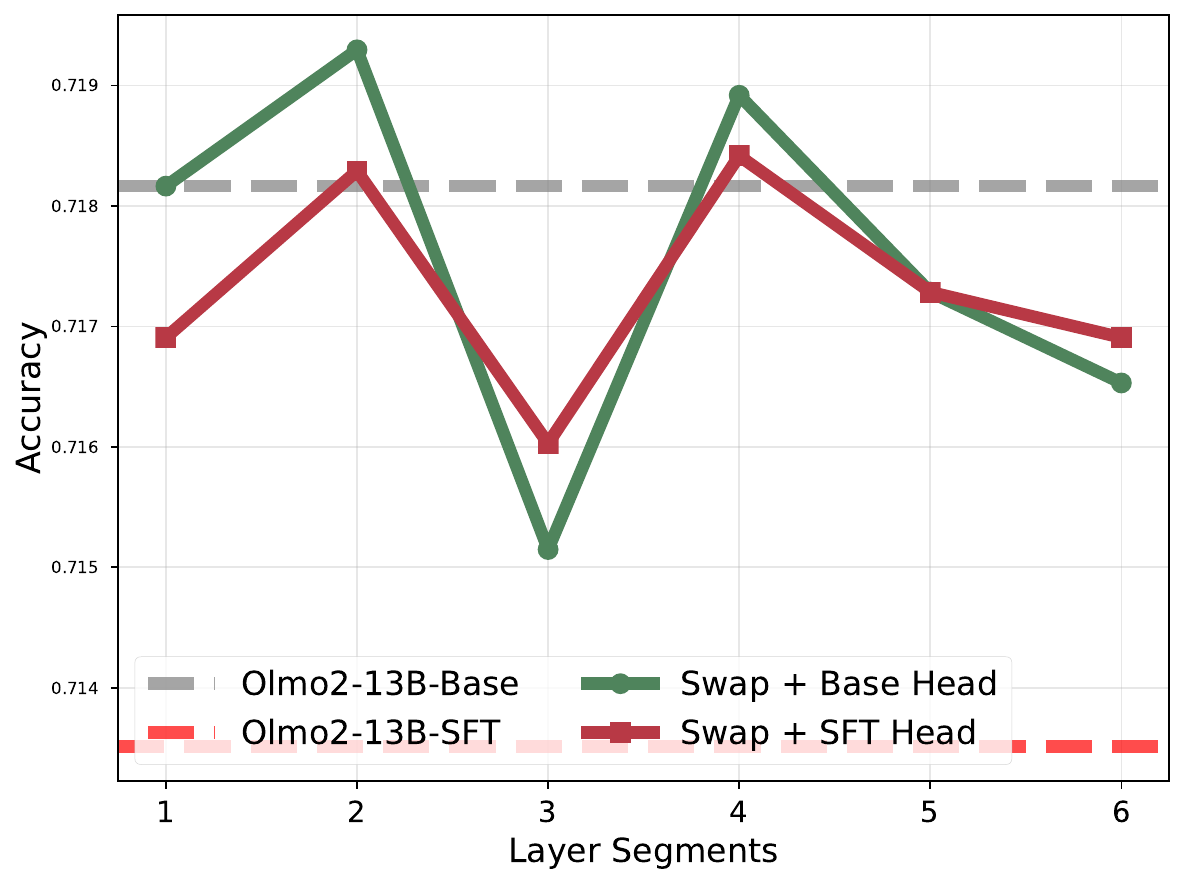}
    \includegraphics[width=0.45\linewidth]{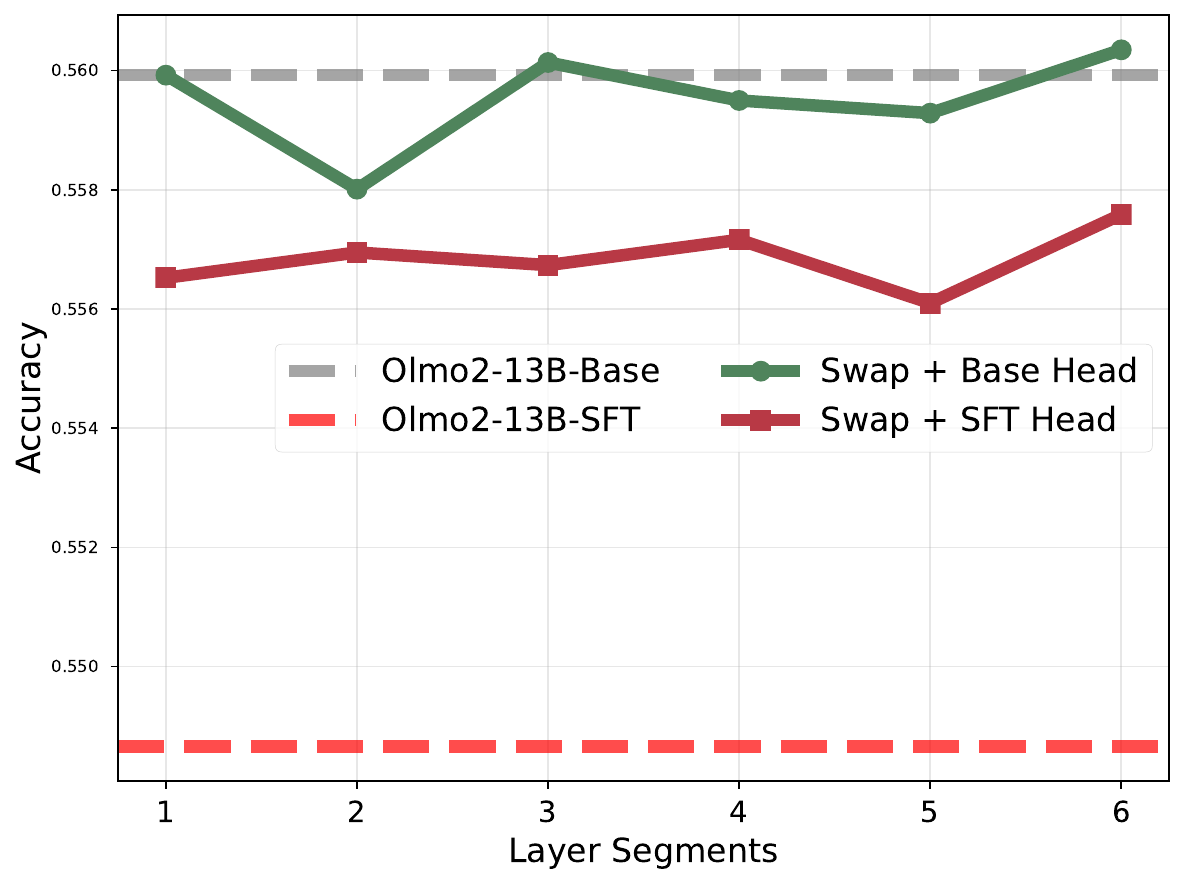}
    \includegraphics[width=0.45\linewidth]{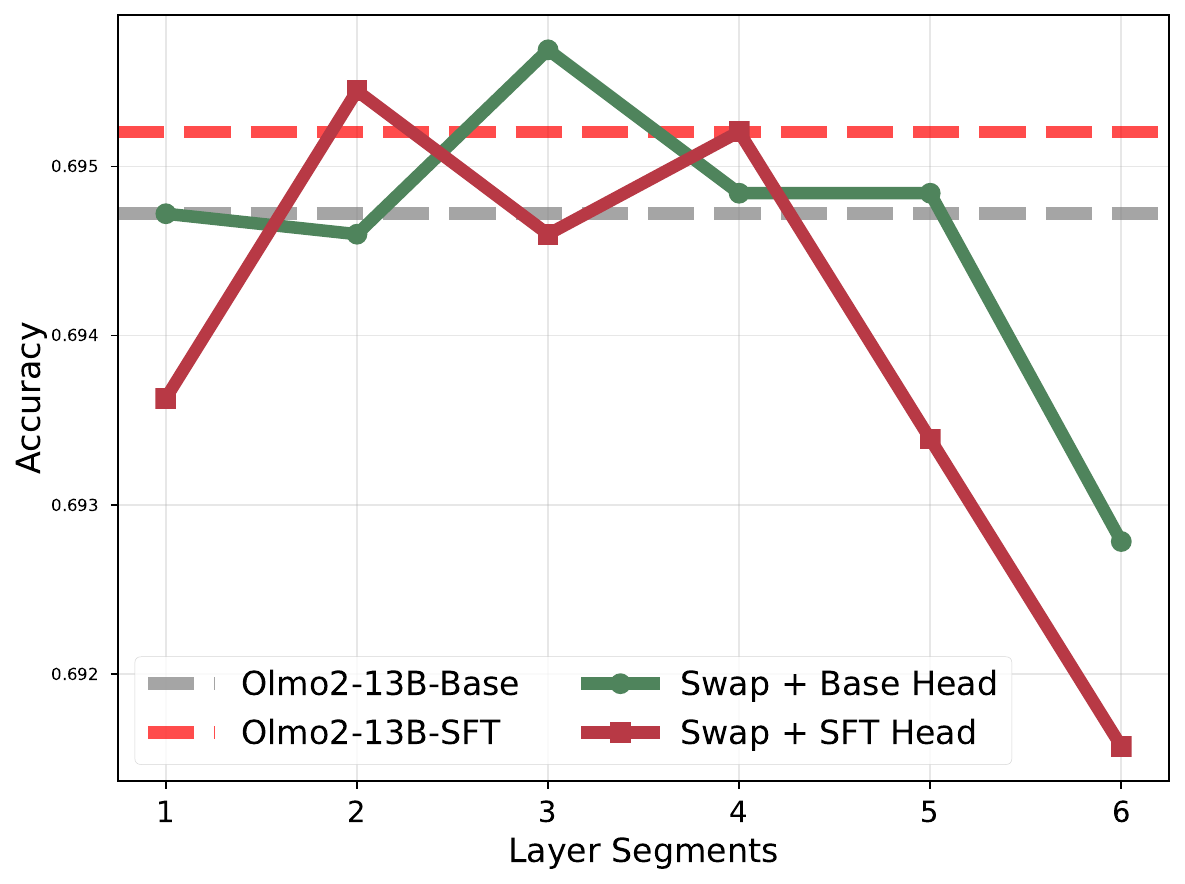}
    \includegraphics[width=0.45\linewidth]{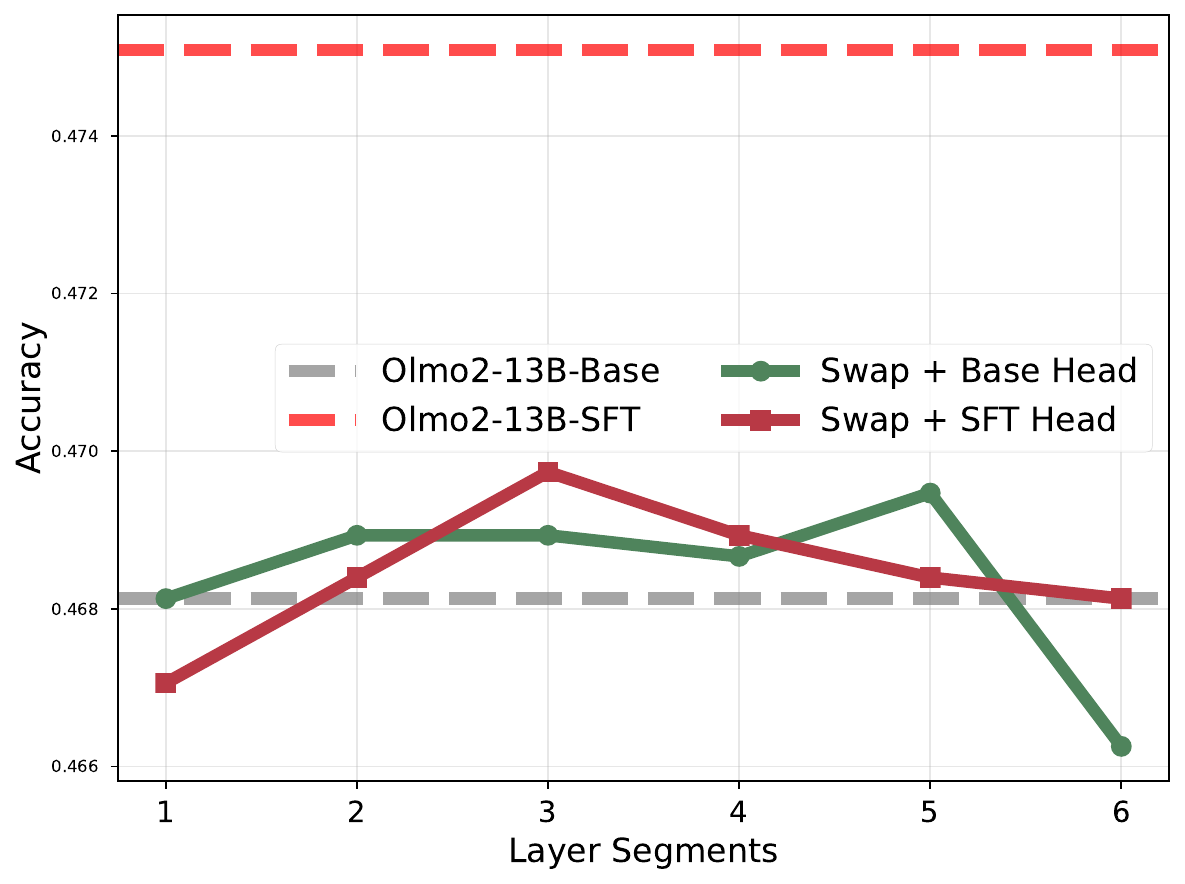}
    \includegraphics[width=0.45\linewidth]{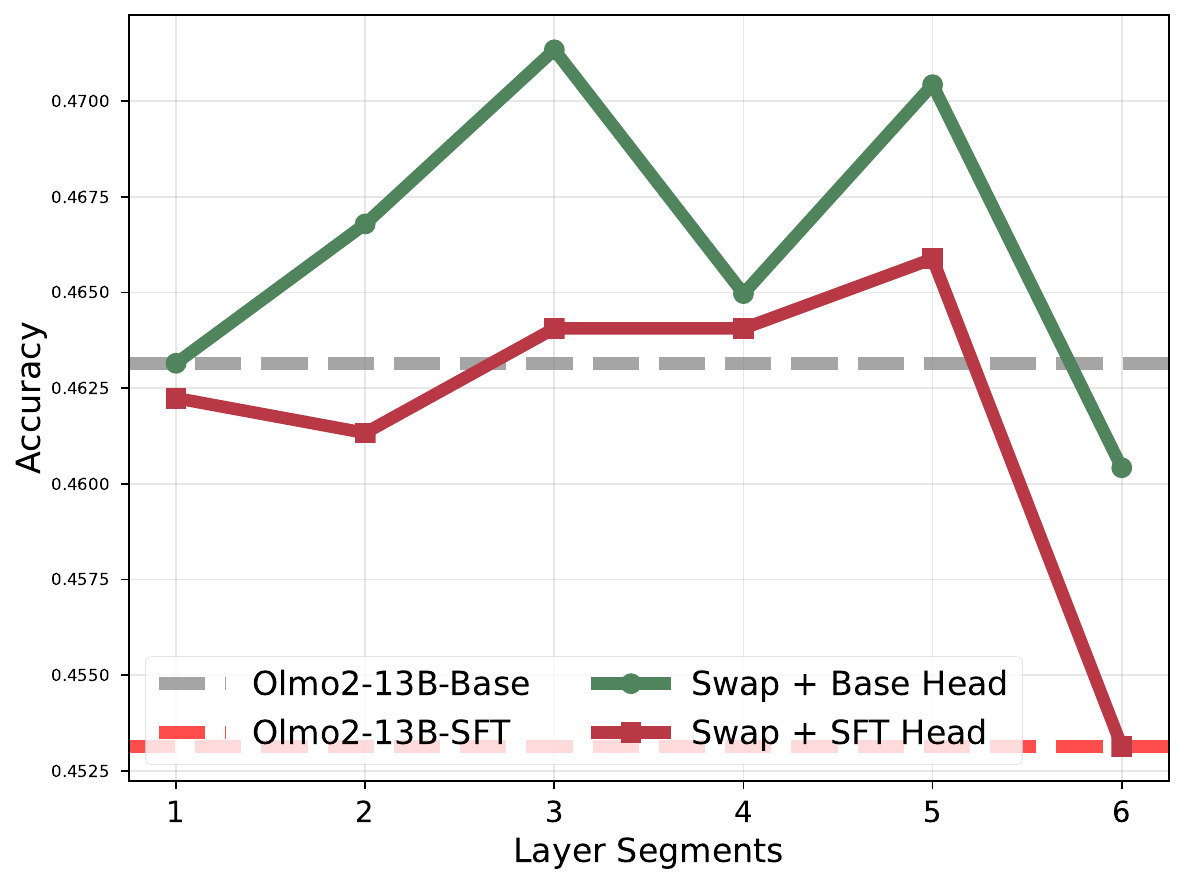}
    \caption{Model performance evaluation under block-wise layer swapping between Base and SFT weights, from left to right is MMLU, GSM8K, WikiText, HumanEval, MT-Bench, ToxiGen dataset.}
    \label{fig:swapping}
\end{figure}

\paragraph{Layer-wise Swapping}
To establish a causal link between layer groups and model performance, we conduct a layer swapping experiment by replacing specific blocks of layers in the Base model with their SFT counterparts (and vice versa).
Figure \ref{fig:swapping} reveals an inverted-U-shaped pattern: replacing layers at either early or late results in  performance degradation, while replacing middle layers can occasionally lead to slight improvements. For instance, on the MMLU dataset using the base head of OLMo2-13B, replacing the first 20\% or last 20\% of layers causes performance drops of 0.001 and 0.002, respectively, in contrast, replacing middle layers produces a  gain of 0.003. Despite the small magnitude of this improvement, the overall trend remains consistent across various datasets and models. We hypothesize that the marginal nature of these changes arises from the limited representational differences between the Base and SFT models, as illustrated in Figure \ref{fig:curvature}.

This counterintuitive observation prompts us to consider the following explanations: 1) The later layers are more task-specific and exhibit stronger coupling with adjacent layers (as shown in Figure \ref{fig:curvature}, where changes in the later layers are more pronounced). Consequently, indiscriminately replacing them leads to performance drops. In contrast, the middle layers have weaker coupling with the early and late layers, as they store more general knowledge (evidenced by the relatively lower accuracy in the early layers shown in Figure \ref{fig:probing}).
The new knowledge introduced by SFT does not overwrite or displace the existing knowledge (i.e., no catastrophic forgetting occurs), instead, it may fully retained and primarily resides in the middle layers.

\begin{figure}
    \centering

    \includegraphics[width=0.48\linewidth]{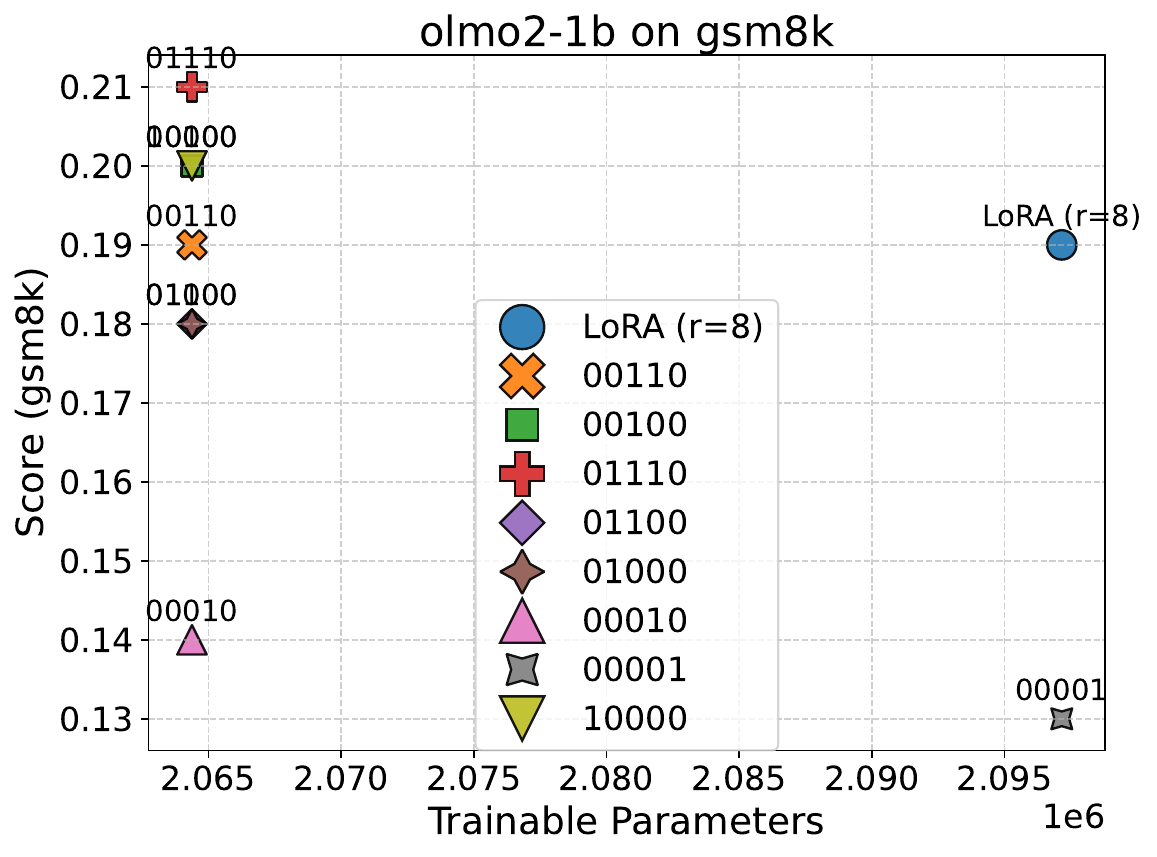}
    \includegraphics[width=0.48\linewidth]{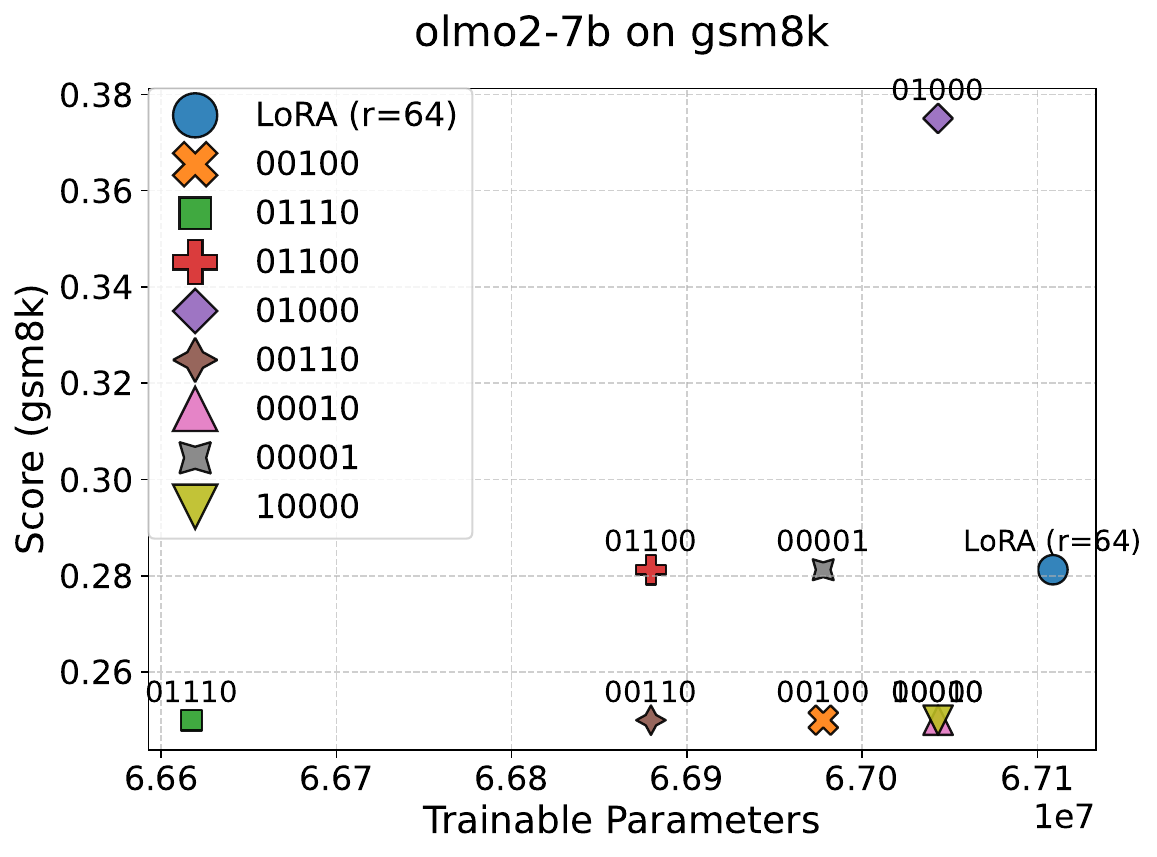}
    \includegraphics[width=0.48\linewidth]{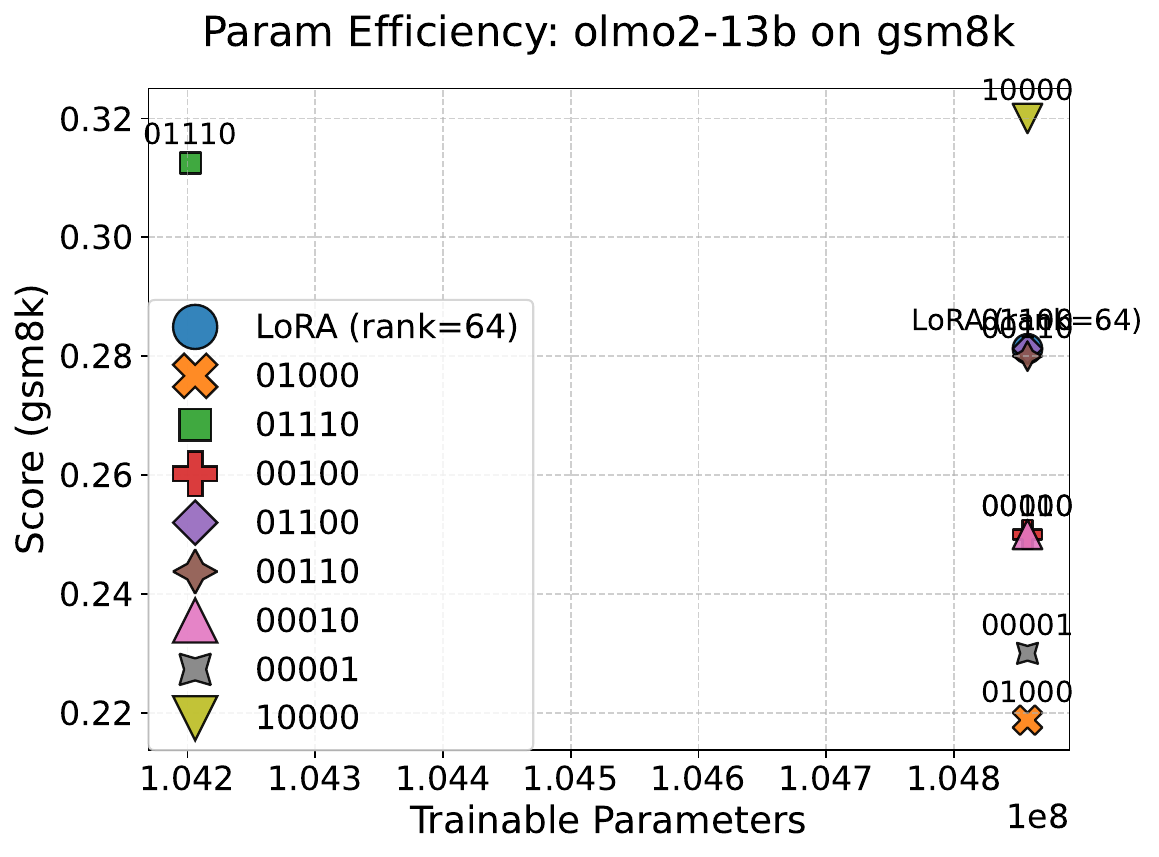}
    \includegraphics[width=0.48\linewidth]{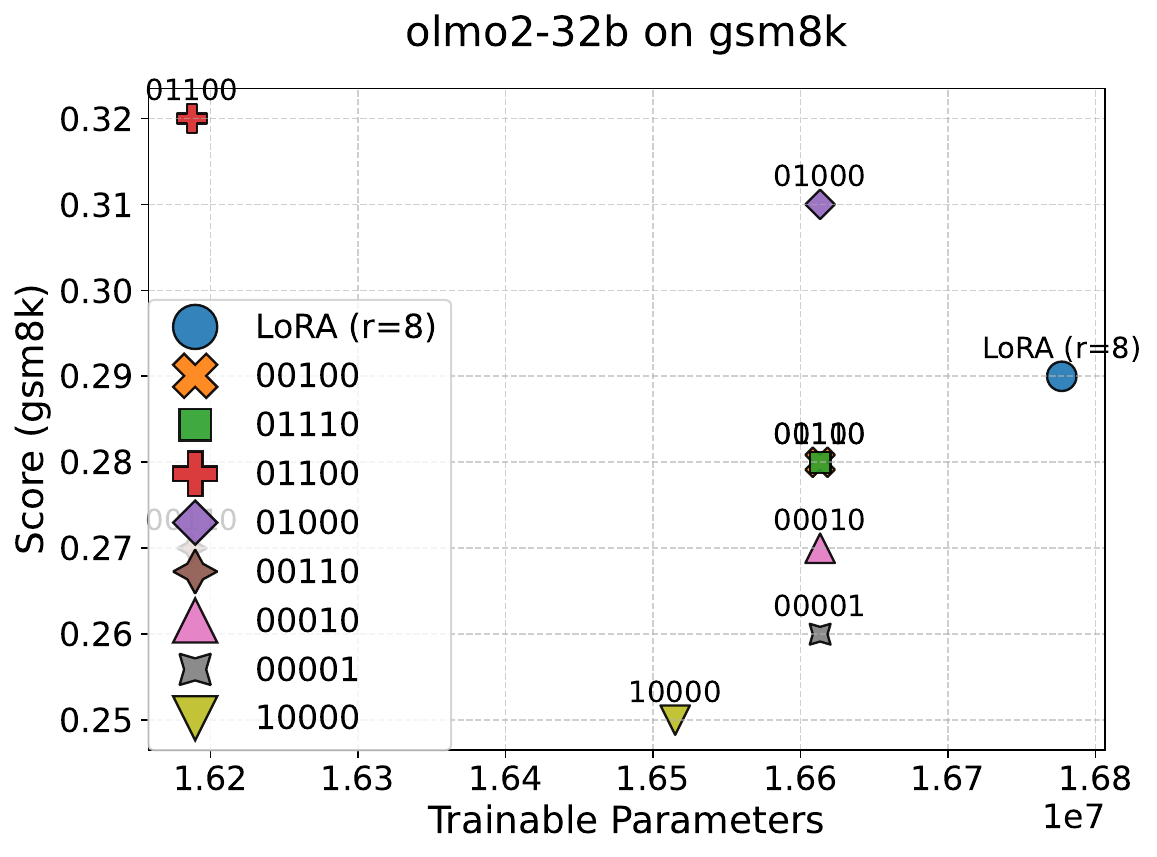}
    \caption{Mid-Block Efficient Tuning, from left to right is OLMo2-(1B, 7B, 13B, 32B), respectively.}
    \label{fig:lora}
\end{figure}
\subsection{Mid-Block Efficient Tuning}\label{sec:efficient_tuning}
To test this, we divide the model layers into $M$ equal segments and compare performance when the added LoRA parameters (with $M=5$ and rank $\in \{8, 64\}$) are concentrated in different segments, while keeping the total number of trainable parameters nearly constant. Experiments are conducted on the OLMo2 family using the GSM8K and MMLU dataset, with GSM8K results shown in Figure \ref{fig:lora}.

For OLMo2-7B, targeting only the upper-middle segment (\texttt{01000}) yielded {0.375}, substantially outperforming the full-layer LoRA baseline (0.28) by nearly {10 percentage points}.  Similarly, on {OLMo2-32B}, the mid-upper configuration (\texttt{01100}) achieved {0.32}, surpassing the baseline (0.29) while using fewer trainable parameters. On {OLMo2-13B}, the wider middle configuration (\texttt{01110}) led with {0.30} versus the baseline's {0.27}.  In contrast, focusing solely on the {bottom segment} (\texttt{10000}) yielded poor results, for instance, 0.22 on OLMo2-13B (vs. baseline 0.27) and 0.25 on the 32B model (7 points below the best middle-segment setting).
Tuning only the {top segment} (\texttt{00001}) performed equally poorly: on OLMo2-1B, it scores just {0.135} (vs. baseline 0.19), indicating that adapting only the output mapping without modifying internal reasoning is insufficient.
Overall, the performance gap between the best middle-segment and worst edge-segment configurations frequently exceeds {20\%} of the total score, underscoring the critical importance of targeted layer selection in LoRA adaptation.

To further investigate the sensitivity of performance to finer-grained boundary choices, we conduct additional ablation studies with $M = 3$ and $M = 10$ segments. The results consistently exhibit an inverted-U-shaped pattern across granularities, confirming that the intermediate region represents a broad and robust representational plateau rather than a sensitive peak. Details are provided in Appendix~\ref{appendix:sensitivity}.
We further  provide quantitative evidence for pre-training knowledge retention as detailed  in  Appendix~\ref{appendix:retention}.

\section*{Conclusion}
By probing the layer-wise dynamics of SFT from information-theoretic, geometric, and optimization perspectives, we identified a consistent depth-dependent adaptation pattern: alignment is not uniformly distributed but architecturally localized. Specifically, we characterize the top layers as sites of aggressive plasticity and information overwriting, whereas middle layers facilitate robust knowledge integration. Guided by these findings, our proposed {Mid-Block Efficient Tuning} exploits this stable intermediate zone, achieving substantial gains over full-depth approaches (e.g., +10.2\% on GSM8K). Our findings suggest that future alignment strategies must move beyond uniform updates to prioritize the functional distinctiveness of model layers, balancing plasticity with stability to mitigate catastrophic forgetting.

\section*{Limitations}

Our study focuses on establishing a mechanistic baseline for layer-wise alignment dynamics, however, we define the following scopes to ensure the precision of our claims, which point toward avenues for future generalization.

First, regarding {architectural scope}, we prioritized our analysis on standard dense decoder-only architectures (OLMo2 and Mistral families) to ensure a controlled experimental environment. While this enables us to isolate the effects of depth without the confounding routing noise of Mixture-of-Experts (MoE) or encoder-decoder interactions, extending our Mid-Block hypothesis to these complex topologies remains a promising direction for future research to verify cross-architecture universality.

Second, we explicitly target the  SFT stage, isolating it from subsequent alignment phases like RLHF or DPO. This scoping is deliberate: it allows us to pinpoint the emergence of instruction-following capabilities at the representational level before preference optimization introduces reward-driven feature shifts. While our findings provide a solid foundation, investigating how these layer-wise dynamics evolve under preference-based objectives will be a valuable extension of this work.

Finally, while our proposed Mid-Block Efficient Tuning demonstrates significant gains by updating the 20\%-80\% depth range, this block selection is currently empirical. Our results indicate that this sweet spot is a broad, robust plateau rather than a sharp, sensitive peak, suggesting high transferability. However, developing an adaptive, training-free metric to automatically identify these optimal boundaries for unseen architectures could further streamline the deployment of this method.

\section*{Acknowledgments}

This work was partially supported by the Hefei College Talent Research Fund Project (No. 24RC20), the Scientific Research Project of the Anhui Provincial Education Department (No. 2025AHGXZK40379), and the Natural Science Research Project of the Anhui Educational Committee (No. 2024AH040209). 
We also thank the anonymous reviewers for their constructive comments and suggestions.

\bibliography{tacl2021}

@article{lima2023,
  title={LIMA: Less Is More for Alignment},
  author={Zhou, Chunting and Liu, Pengfei and Xu, Puxin and Iyer, Srini and Sun, Jiao and Mao, Yuning and Ma, Xuezhe and Efrat, Avia and Yu, Ping and Yu, Lili and others},
  journal={arXiv preprint arXiv:2305.11206},
  year={2023},
  url={https://arxiv.org/abs/2305.11206}
}

@misc{jha2023limitinstructiontuningevaluation,
      title={LIMIT: Less Is More for Instruction Tuning Across Evaluation Paradigms}, 
      author={Aditi Jha and Sam Havens and Jeremy Dohmann and Alex Trott and Jacob Portes},
      year={2023},
      eprint={2311.13133},
      archivePrefix={arXiv},
      primaryClass={cs.LG},
      url={https://arxiv.org/abs/2311.13133}, 
}

@inproceedings{ouyang2022training,
  title={Training language models to follow instructions with human feedback},
  author={Ouyang, Long and Wu, Jeffrey and Jiang, Xu and Almeida, Diogo and Wainwright, Carroll and Mishkin, Pamela and Zhang, Chong and Agarwal, Sandhini and Slama, Katarina and Ray, Alex and others},
  booktitle={Advances in Neural Information Processing Systems},
  volume={35},
  pages={27730--27744},
  year={2022}
}

@article{meng2022locating,
  title={Locating and editing factual associations in {GPT}},
  author={Meng, Kevin and Bau, David and Andonian, Alex and Belinkov, Yonatan},
  journal={Advances in Neural Information Processing Systems},
  volume={35},
  pages={17359--17372},
  year={2022}
}

@article{tenney2019bert,
  title={{BERT} rediscovers the classical {NLP} pipeline},
  author={Tenney, Ian and Das, Dipanjan and Pavlick, Ellie},
  journal={arXiv preprint arXiv:1905.05950},
  year={2019}
}

@inproceedings{10.1145/3630106.3658979,
author = {Huang, Saffron and Siddarth, Divya and Lovitt, Liane and Liao, Thomas I. and Durmus, Esin and Tamkin, Alex and Ganguli, Deep},
title = {Collective Constitutional AI: Aligning a Language Model with Public Input},
year = {2024},
isbn = {9798400704505},
publisher = {Association for Computing Machinery},
address = {New York, NY, USA},
url = {https://doi.org/10.1145/3630106.3658979},
doi = {10.1145/3630106.3658979},
booktitle = {Proceedings of the 2024 ACM Conference on Fairness, Accountability, and Transparency},
pages = {1395–1417},
numpages = {23},
location = {Rio de Janeiro, Brazil},
series = {FAccT '24}
}

@inproceedings{
wei2022finetuned,
title={Finetuned Language Models are Zero-Shot Learners},
author={Jason Wei and Maarten Bosma and Vincent Zhao and Kelvin Guu and Adams Wei Yu and Brian Lester and Nan Du and Andrew M. Dai and Quoc V Le},
booktitle={International Conference on Learning Representations},
year={2022},
url={https://openreview.net/forum?id=gEZrGCozdqR}
}

@inproceedings{NEURIPS2023_6dcf277e,
 author = {Liu, Haotian and Li, Chunyuan and Wu, Qingyang and Lee, Yong Jae},
 booktitle = {Advances in Neural Information Processing Systems},
 editor = {A. Oh and T. Naumann and A. Globerson and K. Saenko and M. Hardt and S. Levine},
 pages = {34892--34916},
 publisher = {Curran Associates, Inc.},
 title = {Visual Instruction Tuning},
 url = {https://proceedings.neurips.cc/paper_files/paper/2023/file/6dcf277ea32ce3288914faf369fe6de0-Paper-Conference.pdf},
 volume = {36},
 year = {2023}
}

@inproceedings{10.5555/3692070.3694024,
author = {Tang, Zhengyang and Zhang, Xingxing and Wang, Benyou and Wei, Furu},
title = {MathScale: scaling instruction tuning for mathematical reasoning},
year = {2024},
publisher = {JMLR.org},
booktitle = {Proceedings of the 41st International Conference on Machine Learning},
articleno = {1954},
numpages = {16},
location = {Vienna, Austria},
series = {ICML'24}
}

@article{10.5555/3722577.3722647,
author = {Chung, Hyung Won and Hou, Le and Longpre, Shayne and Zoph, Barret and Tai, Yi and Fedus, William and Li, Yunxuan and Wang, Xuezhi and Dehghani, Mostafa and Brahma, Siddhartha and Webson, Albert and Gu, Shixiang Shane and Dai, Zhuyun and Suzgun, Mirac and Chen, Xinyun and Chowdhery, Aakanksha and Castro-Ros, Alex and Pellat, Marie and Robinson, Kevin and Valter, Dasha and Narang, Sharan and Mishra, Gaurav and Yu, Adams and Zhao, Vincent and Huang, Yanping and Dai, Andrew and Yu, Hongkun and Petrov, Slav and Chi, Ed H. and Dean, Jeff and Devlin, Jacob and Roberts, Adam and Zhou, Denny and Le, Quoc V. and Wei, Jason},
title = {Scaling instruction-finetuned language models},
year = {2024},
issue_date = {January 2024},
publisher = {JMLR.org},
volume = {25},
number = {1},
issn = {1532-4435},
journal = {J. Mach. Learn. Res.},
month = jan,
articleno = {70},
numpages = {53}
}

@article{lyu2024keeping,
  title={Keeping llms aligned after fine-tuning: The crucial role of prompt templates},
  author={Lyu, Kaifeng and Zhao, Haoyu and Gu, Xinran and Yu, Dingli and Goyal, Anirudh and Arora, Sanjeev},
  journal={Advances in Neural Information Processing Systems},
  volume={37},
  pages={118603--118631},
  year={2024}
}

@inproceedings{rimsky-etal-2024-steering,
    title = "Steering Llama 2 via Contrastive Activation Addition",
    author = "Rimsky, Nina  and
      Gabrieli, Nick  and
      Schulz, Julian  and
      Tong, Meg  and
      Hubinger, Evan  and
      Turner, Alexander",
    editor = "Ku, Lun-Wei  and
      Martins, Andre  and
      Srikumar, Vivek",
    booktitle = "Proceedings of the 62nd Annual Meeting of the Association for Computational Linguistics (Volume 1: Long Papers)",
    month = aug,
    year = "2024",
    address = "Bangkok, Thailand",
    publisher = "Association for Computational Linguistics",
    url = "https://aclanthology.org/2024.acl-long.828/",
    doi = "10.18653/v1/2024.acl-long.828",
    pages = "15504--15522"
}

@inproceedings{wu-etal-2024-language,
    title = "From Language Modeling to Instruction Following: Understanding the Behavior Shift in {LLM}s after Instruction Tuning",
    author = "Wu, Xuansheng  and
      Yao, Wenlin  and
      Chen, Jianshu  and
      Pan, Xiaoman  and
      Wang, Xiaoyang  and
      Liu, Ninghao  and
      Yu, Dong",
    editor = "Duh, Kevin  and
      Gomez, Helena  and
      Bethard, Steven",
    booktitle = "Proceedings of the 2024 Conference of the North American Chapter of the Association for Computational Linguistics: Human Language Technologies (Volume 1: Long Papers)",
    month = jun,
    year = "2024",
    address = "Mexico City, Mexico",
    publisher = "Association for Computational Linguistics",
    url = "https://aclanthology.org/2024.naacl-long.130/",
    doi = "10.18653/v1/2024.naacl-long.130",
    pages = "2341--2369"
}

@inproceedings{
ren2025learning,
title={Learning Dynamics of {LLM} Finetuning},
author={Yi Ren and Danica J. Sutherland},
booktitle={The Thirteenth International Conference on Learning Representations},
year={2025},
url={https://openreview.net/forum?id=tPNHOoZFl9}
}

@misc{du2025posttrainingreshapesllmsmechanistic,
      title={How Post-Training Reshapes LLMs: A Mechanistic View on Knowledge, Truthfulness, Refusal, and Confidence}, 
      author={Hongzhe Du and Weikai Li and Min Cai and Karim Saraipour and Zimin Zhang and Himabindu Lakkaraju and Yizhou Sun and Shichang Zhang},
      year={2025},
      eprint={2504.02904},
      archivePrefix={arXiv},
      primaryClass={cs.CL},
      url={https://arxiv.org/abs/2504.02904}, 
}

@misc{he2025saifsparseautoencoderframework,
      title={SAIF: A Sparse Autoencoder Framework for Interpreting and Steering Instruction Following of Language Models}, 
      author={Zirui He and Haiyan Zhao and Yiran Qiao and Fan Yang and Ali Payani and Jing Ma and Mengnan Du},
      year={2025},
      eprint={2502.11356},
      archivePrefix={arXiv},
      primaryClass={cs.LG},
      url={https://arxiv.org/abs/2502.11356}, 
}

@InProceedings{pmlr-v235-ghosh24a,
  title = 	 {A Closer Look at the Limitations of Instruction Tuning},
  author =       {Ghosh, Sreyan and Evuru, Chandra Kiran Reddy and Kumar, Sonal and S, Ramaneswaran and Aneja, Deepali and Jin, Zeyu and Duraiswami, Ramani and Manocha, Dinesh},
  booktitle = 	 {Proceedings of the 41st International Conference on Machine Learning},
  pages = 	 {15559--15589},
  year = 	 {2024},
  editor = 	 {Salakhutdinov, Ruslan and Kolter, Zico and Heller, Katherine and Weller, Adrian and Oliver, Nuria and Scarlett, Jonathan and Berkenkamp, Felix},
  volume = 	 {235},
  series = 	 {Proceedings of Machine Learning Research},
  month = 	 {21--27 Jul},
  publisher =    {PMLR},
  url = 	 {https://proceedings.mlr.press/v235/ghosh24a.html}
}

@inproceedings{
lin2024the,
title={The Unlocking Spell on Base {LLM}s:  Rethinking Alignment via In-Context Learning},
author={Bill Yuchen Lin and Abhilasha Ravichander and Ximing Lu and Nouha Dziri and Melanie Sclar and Khyathi Chandu and Chandra Bhagavatula and Yejin Choi},
booktitle={The Twelfth International Conference on Learning Representations},
year={2024},
url={https://openreview.net/forum?id=wxJ0eXwwda}
}

@inproceedings{kung-peng-2023-models,
    title = "Do Models Really Learn to Follow Instructions? An Empirical Study of Instruction Tuning",
    author = "Kung, Po-Nien  and
      Peng, Nanyun",
    editor = "Rogers, Anna  and
      Boyd-Graber, Jordan  and
      Okazaki, Naoaki",
    booktitle = "Proceedings of the 61st Annual Meeting of the Association for Computational Linguistics (Volume 2: Short Papers)",
    month = jul,
    year = "2023",
    address = "Toronto, Canada",
    publisher = "Association for Computational Linguistics",
    url = "https://aclanthology.org/2023.acl-short.113/",
    doi = "10.18653/v1/2023.acl-short.113",
    pages = "1317--1328"
}

@inproceedings{ren-etal-2024-learning,
    title = "Learning or Self-aligning? Rethinking Instruction Fine-tuning",
    author = "Ren, Mengjie  and
      Cao, Boxi  and
      Lin, Hongyu  and
      Liu, Cao  and
      Han, Xianpei  and
      Zeng, Ke  and
      Guanglu, Wan  and
      Cai, Xunliang  and
      Sun, Le",
    editor = "Ku, Lun-Wei  and
      Martins, Andre  and
      Srikumar, Vivek",
    booktitle = "Proceedings of the 62nd Annual Meeting of the Association for Computational Linguistics (Volume 1: Long Papers)",
    month = aug,
    year = "2024",
    address = "Bangkok, Thailand",
    publisher = "Association for Computational Linguistics",
    url = "https://aclanthology.org/2024.acl-long.330/",
    doi = "10.18653/v1/2024.acl-long.330",
    pages = "6090--6105"
}

@article{10.1162/tacl_a_00673,
    author = {Itzhak, Itay and Stanovsky, Gabriel and Rosenfeld, Nir and Belinkov, Yonatan},
    title = {Instructed to Bias: Instruction-Tuned Language Models Exhibit Emergent Cognitive Bias},
    journal = {Transactions of the Association for Computational Linguistics},
    volume = {12},
    pages = {771-785},
    year = {2024},
    month = {06},
    issn = {2307-387X},
    doi = {10.1162/tacl_a_00673},
    url = {https://doi.org/10.1162/tacl_a_00673},
    eprint = {https://direct.mit.edu/tacl/article-pdf/doi/10.1162/tacl_a_00673/2377780/tacl_a_00673.pdf},
}

@inproceedings{
jiang2025unlocking,
title={Unlocking the Power of Function Vectors for Characterizing and Mitigating Catastrophic Forgetting in Continual Instruction Tuning},
author={Gangwei Jiang and Caigao JIANG and Zhaoyi Li and Siqiao Xue and JUN ZHOU and Linqi Song and Defu Lian and Ying Wei},
booktitle={The Thirteenth International Conference on Learning Representations},
year={2025},
url={https://openreview.net/forum?id=gc8QAQfXv6}
}

@inproceedings{
aw2024instructiontuning,
title={Instruction-tuning Aligns {LLM}s to the Human Brain},
author={Khai Loong Aw and Syrielle Montariol and Badr AlKhamissi and Martin Schrimpf and Antoine Bosselut},
booktitle={First Conference on Language Modeling},
year={2024},
url={https://openreview.net/forum?id=nXNN0x4wbl}
}

@inproceedings{
hendrycks2021measuring,
title={Measuring Massive Multitask Language Understanding},
author={Dan Hendrycks and Collin Burns and Steven Basart and Andy Zou and Mantas Mazeika and Dawn Song and Jacob Steinhardt},
booktitle={International Conference on Learning Representations},
year={2021},
url={https://openreview.net/forum?id=d7KBjmI3GmQ}
}

@misc{cobbe2021trainingverifierssolvemath,
      title={Training Verifiers to Solve Math Word Problems}, 
      author={Karl Cobbe and Vineet Kosaraju and Mohammad Bavarian and Mark Chen and Heewoo Jun and Lukasz Kaiser and Matthias Plappert and Jerry Tworek and Jacob Hilton and Reiichiro Nakano and Christopher Hesse and John Schulman},
      year={2021},
      eprint={2110.14168},
      archivePrefix={arXiv},
      primaryClass={cs.LG},
      url={https://arxiv.org/abs/2110.14168}, 
}

@inproceedings{
merity2017pointer,
title={Pointer Sentinel Mixture Models},
author={Stephen Merity and Caiming Xiong and James Bradbury and Richard Socher},
booktitle={International Conference on Learning Representations},
year={2017},
url={https://openreview.net/forum?id=Byj72udxe}
}

@misc{zhou2023instructionfollowingevaluationlargelanguage,
      title={Instruction-Following Evaluation for Large Language Models}, 
      author={Jeffrey Zhou and Tianjian Lu and Swaroop Mishra and Siddhartha Brahma and Sujoy Basu and Yi Luan and Denny Zhou and Le Hou},
      year={2023},
      eprint={2311.07911},
      archivePrefix={arXiv},
      primaryClass={cs.CL},
      url={https://arxiv.org/abs/2311.07911}, 
}

@misc{chen2021evaluatinglargelanguagemodels,
      title={Evaluating Large Language Models Trained on Code}, 
      author={Mark Chen and Jerry Tworek and Heewoo Jun and Qiming Yuan and Henrique Ponde de Oliveira Pinto and Jared Kaplan and Harri Edwards and Yuri Burda and Nicholas Joseph and Greg Brockman and Alex Ray and Raul Puri and Gretchen Krueger and Michael Petrov and Heidy Khlaaf and Girish Sastry and Pamela Mishkin and Brooke Chan and Scott Gray and Nick Ryder and Mikhail Pavlov and Alethea Power and Lukasz Kaiser and Mohammad Bavarian and Clemens Winter and Philippe Tillet and Felipe Petroski Such and Dave Cummings and Matthias Plappert and Fotios Chantzis and Elizabeth Barnes and Ariel Herbert-Voss and William Hebgen Guss and Alex Nichol and Alex Paino and Nikolas Tezak and Jie Tang and Igor Babuschkin and Suchir Balaji and Shantanu Jain and William Saunders and Christopher Hesse and Andrew N. Carr and Jan Leike and Josh Achiam and Vedant Misra and Evan Morikawa and Alec Radford and Matthew Knight and Miles Brundage and Mira Murati and Katie Mayer and Peter Welinder and Bob McGrew and Dario Amodei and Sam McCandlish and Ilya Sutskever and Wojciech Zaremba},
      year={2021},
      eprint={2107.03374},
      archivePrefix={arXiv},
      primaryClass={cs.LG},
      url={https://arxiv.org/abs/2107.03374}, 
}

@inproceedings{
zheng2023judging,
title={Judging {LLM}-as-a-Judge with {MT}-Bench and Chatbot Arena},
author={Lianmin Zheng and Wei-Lin Chiang and Ying Sheng and Siyuan Zhuang and Zhanghao Wu and Yonghao Zhuang and Zi Lin and Zhuohan Li and Dacheng Li and Eric Xing and Hao Zhang and Joseph E. Gonzalez and Ion Stoica},
booktitle={Thirty-seventh Conference on Neural Information Processing Systems Datasets and Benchmarks Track},
year={2023},
url={https://openreview.net/forum?id=uccHPGDlao}
}

@inproceedings{hartvigsen-etal-2022-toxigen,
    title = "{T}oxi{G}en: A Large-Scale Machine-Generated Dataset for Adversarial and Implicit Hate Speech Detection",
    author = "Hartvigsen, Thomas  and
      Gabriel, Saadia  and
      Palangi, Hamid  and
      Sap, Maarten  and
      Ray, Dipankar  and
      Kamar, Ece",
    editor = "Muresan, Smaranda  and
      Nakov, Preslav  and
      Villavicencio, Aline",
    booktitle = "Proceedings of the 60th Annual Meeting of the Association for Computational Linguistics (Volume 1: Long Papers)",
    month = may,
    year = "2022",
    address = "Dublin, Ireland",
    publisher = "Association for Computational Linguistics",
    url = "https://aclanthology.org/2022.acl-long.234/",
    doi = "10.18653/v1/2022.acl-long.234",
    pages = "3309--3326"
}

@article{kgqgipm24,
  author       = {Runhao Zhao and
                  Jiuyang Tang and
                  Weixin Zeng and
                  Yunxiao Guo and
                  Xiang Zhao},
  title        = {Towards human-like questioning: Knowledge base question generation
                  with bias-corrected reinforcement learning from human feedback},
  journal      = {Inf. Process. Manag.},
  volume       = {62},
  number       = {3},
  pages        = {104044},
  year         = {2025},
  url          = {https://doi.org/10.1016/j.ipm.2024.104044},
  doi          = {10.1016/J.IPM.2024.104044},
  bibsource    = {dblp computer science bibliography, https://dblp.org}
}

@inproceedings{agenteaicde25,
  author       = {Runhao Zhao and
                  Weixin Zeng and
                  Jiuyang Tang and
                  Yawen Li and
                  Guanhua Ye and
                  Junping Du and
                  Xiang Zhao},
  title        = {Towards Unsupervised Entity Alignment for Highly Heterogeneous Knowledge
                  Graphs},
  booktitle    = {41st {IEEE} International Conference on Data Engineering, {ICDE} 2025,
                  Hong Kong, May 19-23, 2025},
  pages        = {3792--3806},
  publisher    = {{IEEE}},
  year         = {2025},
  url          = {https://doi.org/10.1109/ICDE65448.2025.00283},
  doi          = {10.1109/ICDE65448.2025.00283},
  bibsource    = {dblp computer science bibliography, https://dblp.org}
}

@inproceedings{agentqgcikm24,
  author       = {Runhao Zhao and
                  Jiuyang Tang and
                  Weixin Zeng and
                  Ziyang Chen and
                  Xiang Zhao},
  editor       = {Edoardo Serra and
                  Francesca Spezzano},
  title        = {Zero-shot Knowledge Graph Question Generation via Multi-agent LLMs
                  and Small Models Synthesis},
  booktitle    = {Proceedings of the 33rd {ACM} International Conference on Information
                  and Knowledge Management, {CIKM} 2024, Boise, ID, USA, October 21-25,
                  2024},
  pages        = {3341--3351},
  publisher    = {{ACM}},
  year         = {2024},
  url          = {https://doi.org/10.1145/3627673.3679805},
  doi          = {10.1145/3627673.3679805},
  bibsource    = {dblp computer science bibliography, https://dblp.org}
}

@article{hydra26,
  author       = {Runhao Zhao and
                  Weixin Zeng and
                  Wentao Zhang and
                  Xiang Zhao and
                  Jiuyang Tang and
                  Lei Chen},
  title        = {Towards Temporal Knowledge Graph Alignment in the Wild},
  journal      = {CoRR},
  volume       = {abs/2507.14475},
  year         = {2025},
  url          = {https://doi.org/10.48550/arXiv.2507.14475},
  doi          = {10.48550/ARXIV.2507.14475},
  eprinttype   = {arXiv},
  eprint       = {2507.14475},
  bibsource    = {dblp computer science bibliography, https://dblp.org}
}

\appendix

\section{Sensitivity Analysis of Mid-Block Boundary Choices}
\label{appendix:sensitivity}

To examine whether the performance of Mid-Block Efficient Tuning is sensitive to the choice of segment granularity, we extend our ablation studies by partitioning model layers into $M = 3$ and $M = 10$ equal segments, in addition to the $M = 5$ configuration reported in the main text. We evaluate on OLMo2-13B using the ToxiGen dataset, keeping all other experimental settings identical.

Table~\ref{tab:appendix_m3} reports results for $M = 3$ segments. The middle segment (Segment 2) achieves the highest accuracy (0.4721), outperforming both the first segment (0.4650) and the last segment (0.4550), confirming the inverted-U-shaped trend at a coarse granularity.

\begin{table}[h]
\centering
\caption{Performance with $M = 3$ Segments (OLMo2-13B, ToxiGen).}
\label{tab:appendix_m3}
\begin{tabular}{lc}
\toprule
\textbf{Segment} & \textbf{Accuracy} \\
\midrule
Segment 1 & 0.4650 \\
Segment 2 & \textbf{0.4721} \\
Segment 3 & 0.4550 \\
\bottomrule
\end{tabular}
\end{table}

Table~\ref{tab:appendix_m10} reports results for $M = 10$ segments. Performance rises from Segment 1 (0.4503) to a peak at Segment 5 (0.4771), then declines sharply toward the final segments, with Segment 10 dropping to 0.3821. This finer-grained view reveals that the optimal zone corresponds precisely to the middle layers identified in our main analysis (approximately the 20\%--80\% depth range), and that performance degrades most severely in the tail segments closest to the output.

\begin{table}[h]
\centering
\caption{Performance with $M = 10$ Segments (OLMo2-13B, ToxiGen).}
\label{tab:appendix_m10}
\begin{tabular}{lc}
\toprule
\textbf{Segment} & \textbf{Accuracy} \\
\midrule
Segment 1  & 0.4503 \\
Segment 2  & 0.4589 \\
Segment 3  & 0.4693 \\
Segment 4  & 0.4761 \\
Segment 5  & \textbf{0.4771} \\
Segment 6  & 0.4754 \\
Segment 7  & 0.4502 \\
Segment 8  & 0.4215 \\
Segment 9  & 0.3966 \\
Segment 10 & 0.3821 \\
\bottomrule
\end{tabular}
\end{table}

Taken together, these results demonstrate that the inverted-U-shaped pattern is robust across segment granularities. As the number of segments increases, the pattern becomes more pronounced, with performance peaking in the exact middle layers and sharply degrading toward the output layers. This confirms that the 20\%--80\% interval is not a sensitive heuristic but rather a broad, stable region that is tolerant to minor boundary perturbations.

\begin{table*}[!ht]
\centering
\caption{Summary of instruction-tuned versions of commonly used open-source LLMs.}
\label{tab:appendix_modelsurvey}
\resizebox{\textwidth}{!}{
\begin{tabular}{llll}
\toprule
\textbf{Model Family} & \textbf{Official Aligned Version} & \textbf{Alignment Pipeline} 
& \textbf{Suitable} \\
\midrule
LLaMA 2 / 3  & LLaMA-2-Chat / LLaMA-3-Instruct & SFT + RLHF / DPO  & \xmark \\
Gemma 1 / 2  & Gemma-IT / Gemma-2-IT           & SFT + RLHF        & \xmark \\
Qwen 1.5 / 2 & Qwen-Chat / Qwen2-Instruct      & SFT + DPO         & \xmark \\
DeepSeek     & DeepSeek-LLM-Chat / V2          & SFT + RLHF / DPO  & \xmark \\
Falcon       & Falcon-Instruct                 & Mixed / Unpaired  & \xmark \\
Pythia       & None (Base only)                & N/A               & \xmark \\
OPT          & None (Base only / OPT-IML)      & N/A               & \xmark \\
OLMo2        & OLMo-2-Instruct                 & Pure SFT          & \cmark \\
Mistral      & Mistral-7B-Instruct-v0.1        & Pure SFT          & \cmark \\
\bottomrule
\end{tabular}
}
\end{table*}

\begin{table}[h]
\centering
\caption{Perplexity ($\downarrow$) for pre-training knowledge retention across tuning 
strategies (OLMo2-13B).}
\label{tab:appendix_retention}
\resizebox{\columnwidth}{!}{
\begin{tabular}{lcc}
\toprule
\textbf{Strategy} & \textbf{IFEval} $\downarrow$ & \textbf{MT-Bench} $\downarrow$ \\
\midrule
Base (\texttt{00000})                  & 14.53          & 13.58          \\
Full-layer LoRA (\texttt{11111})       & 14.70          & 13.76          \\
Top-layer Segments (\texttt{10000})    & 14.61          & 13.63          \\
Tail-layer Segments (\texttt{00001})   & 14.65          & 13.72          \\
Middle-layer Segments (\texttt{00110}) & \textbf{14.54} & \textbf{13.60} \\
\bottomrule
\end{tabular}
}
\end{table}

\section{Pre-training Knowledge Retention Analysis}
\label{appendix:retention}

A key motivation for Mid-Block Efficient Tuning is its potential to mitigate catastrophic forgetting by avoiding aggressive updates to the highly plastic final layers. To provide 
direct quantitative evidence for this claim, we measure the retention of pre-trained capabilities across different tuning strategies using perplexity on IFEval and MT-Bench, 
where lower perplexity indicates better preservation of pre-trained knowledge.

We compare five configurations on OLMo2-13B: the untuned Base model (\texttt{00000}), standard full-layer LoRA (\texttt{11111}), Top-layer Segments (\texttt{10000}), Tail-layer Segments (\texttt{00001}), and Middle-layer Segments (\texttt{00110}), following the same segment notation as in Section~4.4.

As shown in Table~\ref{tab:appendix_retention}, Middle-layer Segments achieves perplexity closest to the Base model on both benchmarks (14.54 on IFEval and 13.60 on MT-Bench), substantially outperforming standard full-layer LoRA (14.70 and 13.76) and the tail-layer strategy (14.65 and 13.72). These results demonstrate that concentrating updates on intermediate layers effectively mitigates the catastrophic forgetting induced by standard full-layer LoRA, successfully preserving pre-trained capabilities while maintaining robust task alignment.

\section{Survey of Open-Source LLMs and Alignment Pipelines}
\label{appendix:modelsurvey}

A natural concern is whether our experimental conclusions generalize beyond the two model families studied (OLMo2 and Mistral-7B). We carefully selected these families 
after a thorough survey of other widely-used open-source LLMs, including LLaMA, Gemma, Qwen, DeepSeek, Falcon, Pythia, and OPT. Our investigation reveals that most of these 
models either (1) do not provide a corresponding instruction-tuned version trained exclusively with SFT, or (2) their instruction-tuned versions involve a combination of 
multiple alignment techniques such as SFT, RLHF, and DPO, which introduces confounding factors inconsistent with our experimental setup.

Table~\ref{tab:appendix_modelsurvey} summarizes the alignment pipelines of commonly used open-source LLMs. OLMo2 is selected for its full model family coverage across 
multiple scales (1B, 7B, 13B, 32B), and Mistral-7B is selected for its clean SFT-only instruction-tuned variant. Both are the only families that satisfy our requirement of 
a clean base, SFT model pair trained without additional preference optimization stages.

This survey confirms that OLMo2 and Mistral-7B represent the most appropriate and controlled experimental setting for isolating the effects of supervised 
fine-tuning, and that the limited architectural scope is a deliberate methodological choice rather than an oversight.

\end{document}